\documentclass[]{amap}

\usepackage[toc,page,header]{appendix}
\usepackage{minitoc}
\usepackage{solarized-light}
\usepackage{booktabs}
\usepackage{multirow}
\usepackage{graphicx}
\usepackage{array}
\usepackage{siunitx,array}
\usepackage[table]{xcolor}
\usepackage{makecell,array,booktabs}
\usepackage{makecell}
\usepackage{framed}
\usepackage{amssymb}
\usepackage{bbding}
\usepackage{adjustbox}
\usepackage{hyperref}
\usepackage{amsmath} 
\usepackage{placeins}
\usepackage[colorinlistoftodos]{todonotes}
\usepackage{longtable}
\usepackage{hhline}
\usepackage{fancyvrb}
\usepackage{graphicx}
\usepackage[table]{xcolor}
\usepackage{booktabs}
\usepackage{float}
\usepackage{fvextra}
\usepackage{CJKutf8}
\usepackage{multicol}
\usepackage{float}
\usepackage{placeins}
\usepackage{cleveref}
\usepackage{tablefootnote}
\usepackage{threeparttable}
\usepackage{tabularx}
\usepackage{mdframed}
\usepackage{subcaption}
\usepackage[usestackEOL]{stackengine}
\usepackage[numbers]{natbib}
\newcommand{\commentout}[1]{}
\renewcommand{\paragraph}[1]{\noindent\textbf{#1.}\hspace*{1em}}
\usepackage{enumitem}
\usepackage{hyperref}
\setlist[itemize]{leftmargin=15pt}
\usepackage{siunitx,array}
\RequirePackage{xspace}
\makeatletter
\DeclareRobustCommand\onedot{\futurelet\@let@token\@onedot}
\def\@onedot{\ifx\@let@token.\else.\null\fi\xspace}

\makeatother
\newcommand{\MODEL}{ABot-N0\xspace}

\title{\MODEL: Technical Report on the VLA Foundation Model for Versatile Embodied Navigation}

\author{AMAP CV Lab}
\vspace{-11pt}

\contribution{See \hyperref[sec:contributions]{Contributions and Acknowledgments} section for a full author list.}

\abstract{

Embodied navigation has long been fragmented by task-specific architectures. We introduce \textbf{\MODEL}, a unified Vision-Language-Action (VLA) foundation model that achieves a ``Grand Unification'' across \textbf{5} core tasks: Point-Goal, Object-Goal, Instruction-Following, POI-Goal, and Person-Following. \MODEL utilizes a hierarchical ``Brain-Action'' architecture, pairing an LLM-based Cognitive Brain for semantic reasoning with a Flow Matching-based Action Expert for precise, continuous trajectory generation.

To support large-scale learning, we developed the \textbf{\MODEL Data Engine}, curating \textbf{16.9M} expert trajectories and \textbf{5.0M} reasoning samples across \textbf{7,802} high-fidelity 3D scenes (\textbf{10.7 $\text{km}^2$}). \MODEL achieves new \textbf{SOTA} performance across \textbf{7} benchmarks, significantly outperforming specialized models. Furthermore, our Agentic Navigation System integrates a planner with hierarchical topological memory, enabling robust, long-horizon missions in dynamic real-world environments.

\bigskip

\textbf{Correspondence:} chuzedong.czd@alibaba-inc.com, tenan.xsc@alibaba-inc.com

\textbf{Project Page:} \url{https://amap-cvlab.github.io/ABot-Navigation/ABot-N0/}
}

\begin{document}
\maketitle
\vspace{-4pt}


\begin{figure}[!h]
    \centering
    \vspace{-20pt} 
\includegraphics[width=\linewidth]{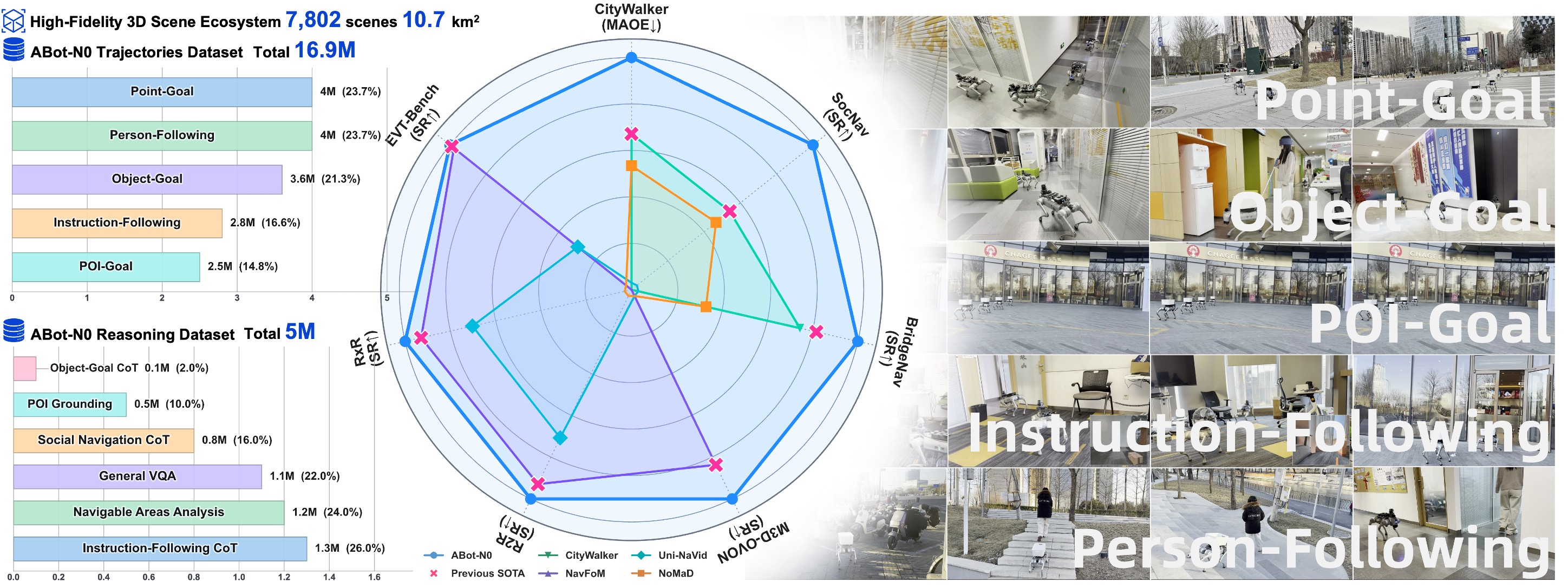}
\caption{\textbf{ABot-N0: A unified VLA foundation model for versatile embodied navigation.} Powered by a massive dataset of 16.9M expert trajectories and 5M reasoning samples across diverse environments, the model achieves a "Grand Unification" across five core navigation tasks. It establishes new state-of-the-art performance across 7 challenging benchmarks and is successfully deployed in complex, dynamic real-world agentic navigation systems.}
\label{fig:model}
\vspace{-20pt} 
\end{figure}

\newpage
\tableofcontents
\newpage


\section{Introduction}

The ultimate objective of Embodied AI is to develop general-purpose agents capable of perceiving complex human intentions and executing autonomous actions within non-structural, open-world environments. In this grand paradigm, navigation serves not only as a fundamental locomotion primitive but also as the critical bridge connecting high-level cognitive reasoning with low-level continuous motor control. However, contemporary navigation research remains deeply entrenched in a ``fragmented paradigm''. Existing methodologies \cite{liu2025citywalker,chen2025socialnav,zhu2025move,xiang2025navr2dualrelationreasoninggeneralizable,yokoyama2024vlfm,wei2025streamvln,long2024instructnav,liu2025navforesee,zhao2026bridgingindooroutdoorgapvisioncentric,wang2025trackvla,liu2025trackvla++, hu2025astranavworldworldmodelforesight} often rely on isolated and specialized architectures tailored for specific tasks—such as Point-Goal, Object-Goal, or Instruction-Following navigation—which inherently limits cross-task generalization and prevents agents from extracting a unified physical prior from large-scale heterogeneous data.

To overcome these limitations, we introduce \textbf{\MODEL}, a unified Vision-Language-Action (VLA) foundation model designed for universal embodied navigation. \MODEL challenges the traditional task-isolation approach by adopting a hierarchical ``Brain-Action'' design philosophy. The architecture comprises three pillars: a \textbf{Universal Multi-Modal Encoder} that unifies heterogeneous inputs into a shared latent space; a \textbf{Cognitive Brain} based on a pre-trained Large Language Model (LLM) for deep semantic understanding and spatial reasoning; and an \textbf{Action Expert} utilizing Flow Matching to generate precise, multi-modal trajectory distributions for continuous control.

Within the \MODEL framework, we achieve a ``Grand Unification'' of five core navigation tasks:
\begin{itemize}
    \item \textbf{Point-Goal Navigation}: The agent must reach precise metric coordinates defined in a local frame, serving as the foundational primitive for robust locomotion and obstacle avoidance \cite{liu2025citywalker, chen2025socialnav}.
    \item \textbf{Object-Goal Navigation}: The agent actively searches for and navigates to a specific object category in unseen environments, requiring sophisticated semantic reasoning and multimodal integration \cite{zhu2025move, xiang2025navr2dualrelationreasoninggeneralizable,yokoyama2024vlfm}.
    \item \textbf{Instruction-Following Navigation}: The agent must execute long-horizon, complex natural language paths, focusing on the rigorous alignment between linguistic input and sequential action execution \cite{wei2025streamvln, long2024instructnav,liu2025navforesee}.
    \item \textbf{POI-Goal Navigation}: This task is defined by \cite{zhao2026bridgingindooroutdoorgapvisioncentric}, which requires the agent to identify specific Points of Interest (POIs) and precisely navigate to their physical entrances, bridging outdoor and indoor environments while addressing the last-meters navigation challenge.
    \item \textbf{Person-Following Navigation}: This involves real-time tracking of dynamic human targets, representing a critical social capability for human-robot interaction \cite{wang2025trackvla, liu2025trackvla++}.
\end{itemize}

The core contributions of this work are summarized in three dimensions:

\paragraph{1. A Unified Embodied Foundation Model Achieving Comprehensive SOTA}
\MODEL represents the first foundation model to achieve a ``Grand Unification'' across five core navigation tasks within a single architecture. Our evaluations demonstrate that \MODEL sets new performance benchmarks (SOTA) across a wide range of authoritative platforms, including CityWalker \cite{liu2025citywalker}, SocNav \cite{chen2025socialnav}, VLN-CE (R2R/RxR) \cite{krantz_vlnce_2020, ku2020room}, HM3D-OVON \cite{yokoyama2024hm3d}, BridgeNav \cite{zhao2026bridgingindooroutdoorgapvisioncentric}, and EVT-Bench \cite{wang2025trackvla}.

\paragraph{2. The Largest Embodied Navigation Data Engine}
The generalization capability of \MODEL is anchored by the \MODEL Data Engine, which has synthesized a standardized corpus of unprecedented scale. This includes a High-Fidelity 3D Scene Ecosystem comprising \textbf{7,802 scenes} covering \textbf{10.7 $\bm{km^2}$} of the total area ($6.25\ km^2$ indoor and $4.42\ km^2$ outdoor). Leveraging these assets, we curated approximately \textbf{16.9 million} expert training trajectories and \textbf{5.0 million} cognitive reasoning samples to enable robust and generalizable navigation skills.

\paragraph{3. A Deployable Agentic Navigation System}
To bridge the gap between foundation models and practical utility, we propose the Agentic Navigation System. This framework augments \MODEL with an \textbf{Agentic Planner} that utilizes Chain-of-Thought (CoT) reasoning for intent decomposition and closed-loop error recovery, a hierarchical \textbf{Topo-Memory (Map-as-Memory)} for cross-scale spatial knowledge deposition. The system has been successfully deployed on the \textbf{Unitree Go2} quadrupedal robot using an \textbf{NVIDIA Jetson Orin NX} module, achieving 2Hz VLA inference and 10Hz closed-loop control in complex, dynamic real-world environments.

\section{\MODEL}
\label{sec:foundation_model}

Existing navigation approaches often treat distinct tasks as separate domains requiring specialized architectures. We challenge this fragmented paradigm by introducing \textbf{\MODEL}, a unified Vision-Language-Action foundation model. \MODEL is designed to combine high-level cognitive reasoning with low-level motion planning, seamlessly generalizing across these five core navigation tasks.

\begin{figure*}[t]
    \centering
    \includegraphics[width=\linewidth]{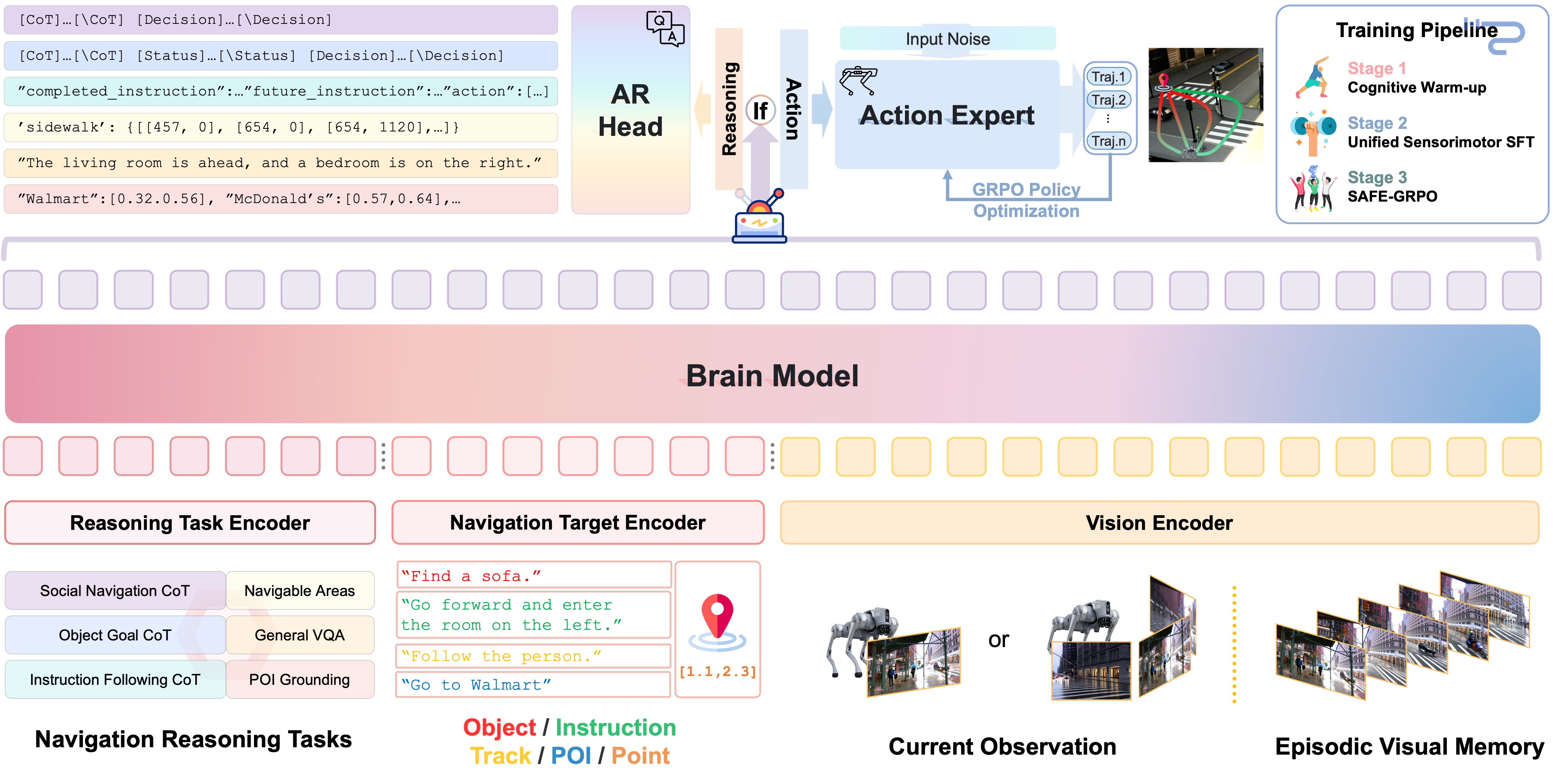}
    \caption{\textbf{The Architecture of \MODEL.} The model adopts a hierarchical ``Brain-Action'' design. The Universal Multi-Modal Encoder unifies heterogeneous inputs (RGB observations, visual history, and goal specifications) into a shared token sequence. The Cognitive Brain (i.e., LLM) encodes these tokens and supports dual-mode operation: a Reasoning Head for high-level semantic understanding and an Action Head for motion planning. The Action Expert employs Flow Matching to generate trajectory distributions, enabling generalization across five navigation tasks.}
    \label{fig:model_arch}
\end{figure*}

The architecture follows a hierarchical ``Brain-Action'' design philosophy (inspired by SocialNav \cite{chen2025socialnav}), comprising three key pillars (See Fig.\ref{fig:model_arch}): 
\begin{itemize}
    \item A \textbf{Universal Multi-Modal Encoder} that unifies heterogeneous inputs into a shared latent space;
    \item A \textbf{Cognitive Brain} based on an Auto-Regressive Large Language Model that encodes generalizable semantic understanding;
    \item An \textbf{Action Expert} that uses Flow Matching \cite{lipman2022flow} to generate precise, multi-modal trajectory distributions.
\end{itemize}

\subsection{Universal Multi-Modal Encoder}

To achieve a ``Grand Unification'' of embodied navigation tasks, \MODEL processes diverse sensor configurations and goal specifications through a flexible token-based encoding scheme.

\subsubsection{Flexible Vision Interface}
\MODEL supports agnostic visual inputs to accommodate different robot morphologies:

\begin{itemize}
    \item \textbf{Current Observation ($O_t$):} We utilize a Vision Transformer (ViT) \cite{dosovitskiy2020image} to encode RGB observations. The interface supports dynamic configurations. In the panoramic mode, the left, front and right views are fed individually into the vision encoder, distinguished by view-specific special tokens. This design preserves the spatial semantics of each view without the geometric distortion introduced by image stitching. In the front-view mode, only the single forward-facing image is processed.
    
    \item \textbf{Episodic Visual Memory ($M_{S}$):} \MODEL maintains an explicit visual history buffer. Relevant historical frames are encoded and appended to the context window, allowing the model to handle Partially Observable Markov Decision Processes (POMDPs).
\end{itemize}

\subsubsection{Heterogeneous Navigation Target Encoder}
\MODEL is able to interpret disparate goal definitions through a unified interface:

\begin{itemize}
    \item \textbf{Semantic Goals (Text-based):} This category encompasses Instruction-Following, Object-Goal, POI-Goal, and Person-Following tasks. Whether the goal is a natural language command, a specific object category, a Point of Interest (POI) name, or a textual description of a target person, we employ the text tokenizer of LLM to embed the goal description directly.
    
    \item \textbf{Geometric Goals (Point-based):} For Point-Goal tasks, the target coordinates $(x, y)$ are defined in the local Bird's Eye View (BEV) frame. These coordinates are projected into the shared embedding dimension via an MLP, allowing the LLM to process geometric vectors as pseudo-tokens.
\end{itemize}

\subsubsection{Reasoning Task Encoder}
To bridge the gap between perception and action, we introduce a Reasoning Task Encoder. This module injects specific task descriptions to activate relevant reasoning circuits within the LLM (e.g., \textit{``Where is Luckin Coffee?''} or \textit{``Identify the zebra crossing area''}). These tasks serve as auxiliary objectives during training, helping the Brain build a robust representation of the environment.

\subsection{The Cognitive Brain}

The backbone of \MODEL is a pre-trained LLM (Qwen3-4B \cite{yang2025qwen3}). It takes the reasoning instructions, navigation goals, visual history, and current observations as input. 

Unlike sequential CoT approaches used in recent LLMs \cite{guo2025deepseek,yang2025qwen3}, our architecture employs a task-conditional design. The Reasoning Head and the Action Head operate as distinct branches (an ``If-Else'' relationship based on the task token), rather than a strictly serial pipeline during inference:

\begin{itemize}
    \item \textbf{Cognitive Activation:} During training, the model is supervised to perform explicit reasoning tasks—such as analyzing scene traversability, identifying social norms, or grounding distinct POIs. These auxiliary tasks are crucial for aligning the high-level semantic representation with the constraints of the physical world .
    \item \textbf{Navigation Decision Support:} For navigation tasks, the Brain leverages the rich, physically-grounded latent context cultivated by the reasoning tasks to directly condition the Action Expert.
\end{itemize}

\subsection{The Action Expert}

For precise motion control, we employ a Flow Matching Action Head, conditional on the context provided by the Brain. The Action Expert predicts a short-term trajectory plan consisting of a sequence of 5 waypoints in the local BEV frame. To ensure full control over the robot's pose, each waypoint includes both 2D coordinates and the yaw orientation:
\begin{equation}
    \mathcal{W} = \{(x_1, y_1, \theta_1), (x_2, y_2, \theta_2), \dots, (x_5, y_5, \theta_5)\}
\end{equation}
where $(x_i, y_i)$ denotes the position and $\theta_i$ represents the local heading angle at step $i$.

We adopt Flow Matching over deterministic regression for two primary reasons:
\begin{itemize}
    \item \textbf{Continuous Precision:} It enables the generation of high-precision continuous values for both translation and rotation, which are essential for smooth robot control and stable heading adjustments.
    \item \textbf{Multi-Modal Distribution Modeling:} In large-scale navigation datasets, similar input conditions often correspond to multiple valid expert behaviors (e.g., bypassing an obstacle from either the left or the right). Deterministic regression tends to average these divergent modes, leading to invalid or collision-prone paths. Flow Matching effectively models this complex distribution, allowing the planner to sample valid, distinct trajectories that reflect the diversity of expert demonstrations.
\end{itemize}
\section{Data Engine}

To bridge the critical gap between perceptual understanding and actionable policy, we introduce the \textbf{\MODEL Data Engine}, a unified synthesis pipeline designed for scalable embodied navigation learning. This engine integrates three synergistic layers: a \textbf{High-Fidelity 3D Scene Ecosystem} (Sec.~\ref{sec:data_source}) offering photorealistic, semantically annotated environments; a \textbf{Universal Trajectories Dataset} (Sec.~\ref{sec:trajectory_data}) aggregating expert demonstrations across five core navigation paradigms (Point-Goal, Object-Goal, Instruction-Goal, POI-Goal, and Person-Following); and a \textbf{Cognitive Reasoning Dataset} (Sec.~\ref{sec:reasoning_dataset}) that grounds decision-making in explicit spatial-social logic. By systematically aligning high-level intent with precise execution, this pipeline provides the essential data fuel for training a generalist, socially compliant navigator.

\begin{figure*}[t]
    \centering
    \includegraphics[width=\linewidth]{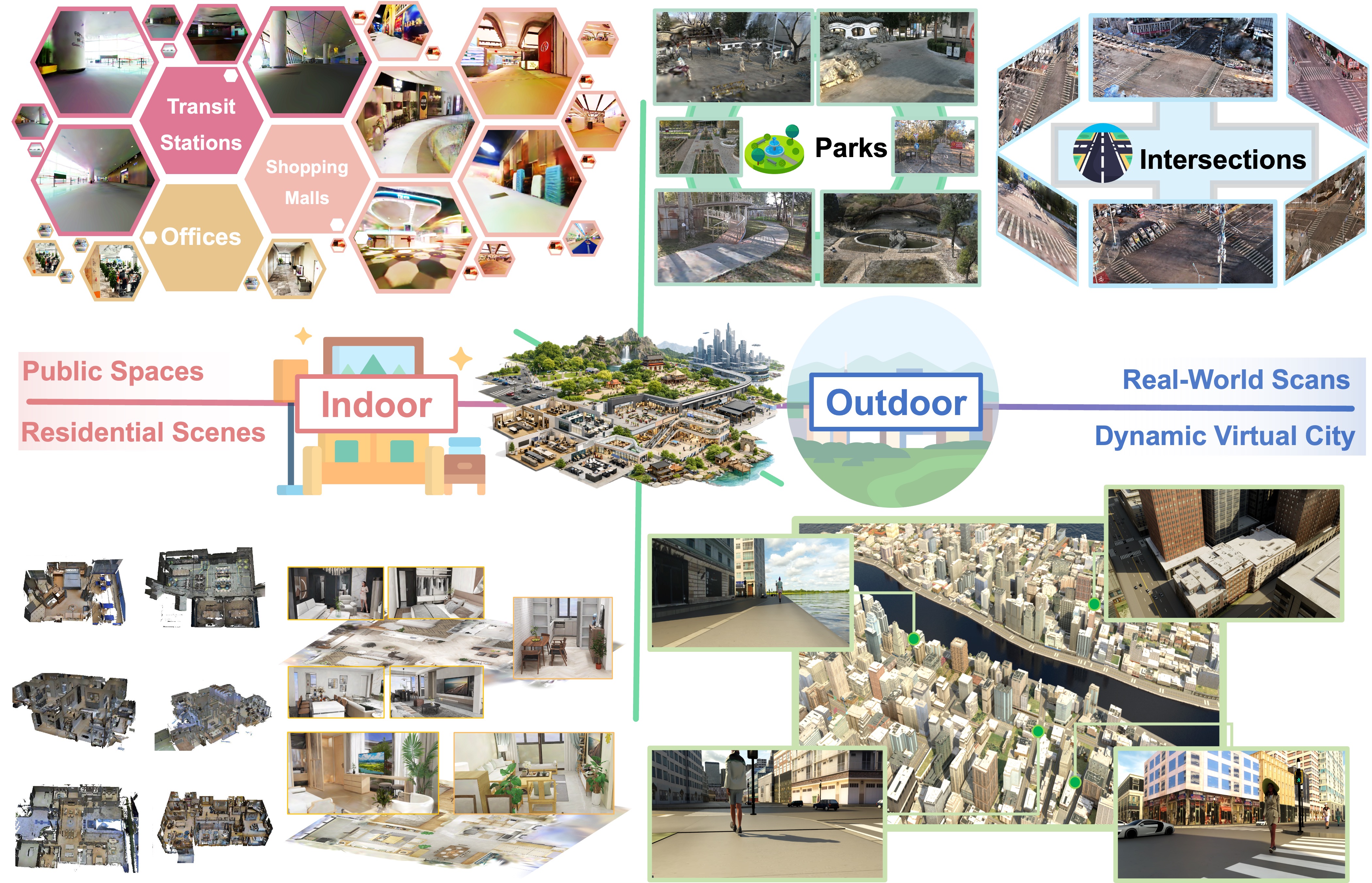}
    \caption{\textbf{High-Fidelity 3D Scene Ecosystem.} Our data engine integrates diverse indoor and outdoor environments. \textit{Left}: Indoor scenes span residential spaces to large-scale public venues (offices, malls, transit stations). \textit{Right}: Outdoor scenes include real-world scans of intersections and parks, alongside the dynamic virtual city SocCity. All scenes are annotated with traversable navigation graphs for collision-free trajectories generation.}
    \label{fig:scene}
\end{figure*}

\subsection{High-Fidelity 3D Scene Ecosystem}
\label{sec:data_source}

\begin{figure*}[t]
    \centering
    \includegraphics[width=0.75\linewidth]{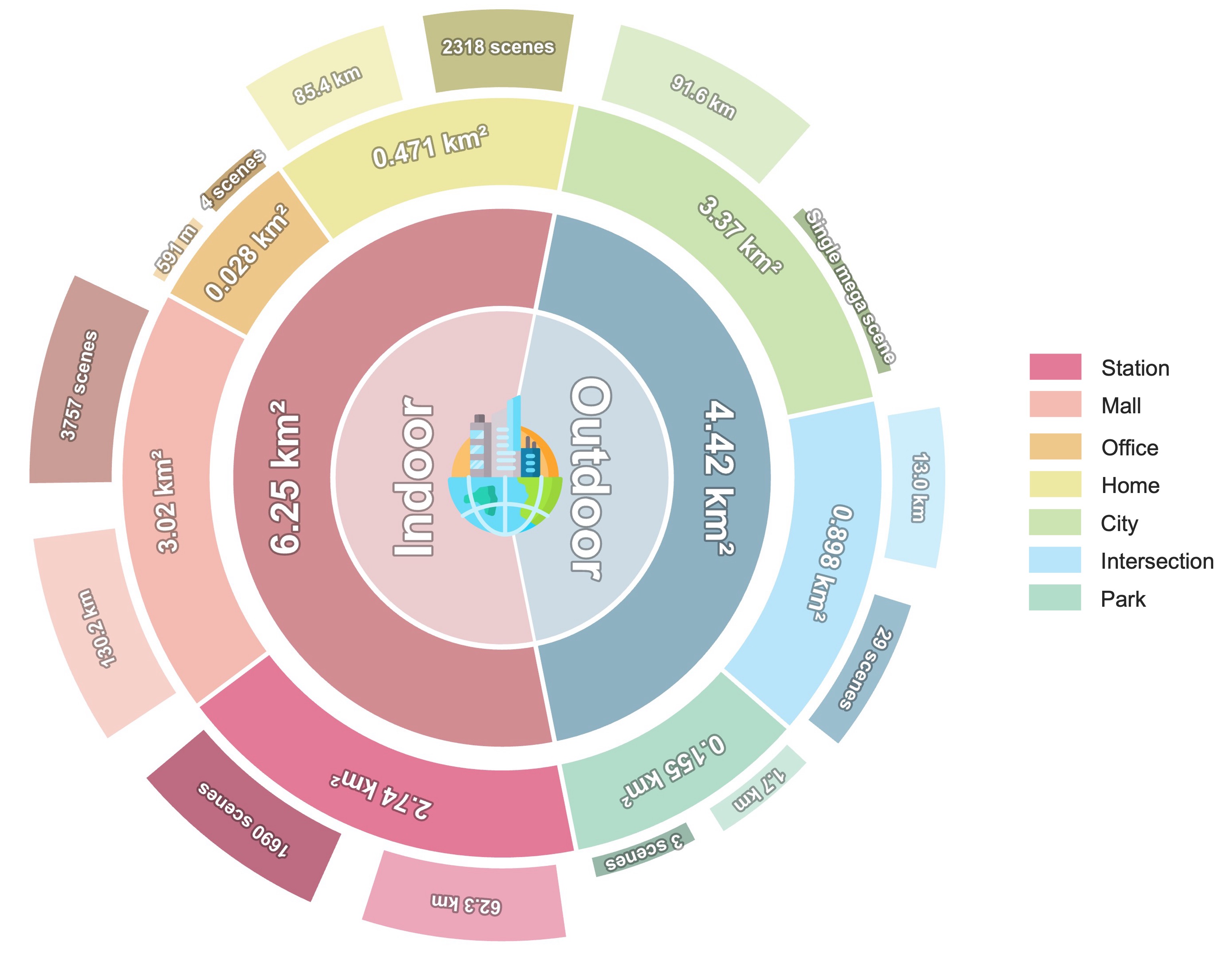}
    \caption{\textbf{3D Scene Ecosystem Statistics.} Our collection comprises \textbf{7,802} high-fidelity 3D scenes, covering \textbf{6.25 $\bm{km^2}$} of indoor environments (offices, malls, stations, homes) and \textbf{4.42 $\bm{km^2}$} of outdoor environments (intersections, parks, city). Navigation graphs totaling \textbf{384,754 meters} enable collision-free trajectory synthesis across diverse spatial scales—from compact residential units to expansive transit hubs and urban environments.}
    \label{fig:scene_stats}
\end{figure*}

To construct a foundation model capable of generalizing across diverse physical environments, we collect and build high-fidelity 3D scenes to support the closed-loop simulation. As shown in Fig.\ref{fig:scene}, we categorize our environments into \textbf{Indoor} and \textbf{Outdoor} domains, integrating widely-used academic benchmarks with our large-scale proprietary 3D Gaussian Splatting (3DGS) scans \cite{kerbl20233dgs}. Detailed statistics of our 3D scene collection are summarized in Fig.\ref{fig:scene_stats}.

Crucially, we implement a unified pipeline for navigation graph annotation to transform these raw visual reconstructions into navigable environments. Every scene in our collection has been manually annotated with traversable road networks, ensuring that all generated expert trajectories are collision-free and socially compliant.

\subsubsection{Indoor Environments: From Domestic to Public}
Our indoor collection is designed to bridge the gap between narrow domestic spaces and unstructured public venues.

\paragraph{Residential Scenes (Home)}
Domestic environments are critical for embodied agents serving household roles. We integrate HM3D \cite{ramakrishnan2021habitat} and InteriorGS \cite{InteriorGS2025}, comprising a total of \textbf{2,318} diverse residential units. These scenes provide rich semantic layouts (e.g., bedrooms, kitchens) essential for Object-Goal and Instruction-Following navigation. The annotated navigation graphs in these scenes strictly define walkable floor space, excluding furniture and obstacles to ensure precise interaction.

\paragraph{Public Spaces (Office, Mall, Station)}
Moving beyond the home, we introduce a large-scale proprietary dataset of public interiors reconstructed via 3DGS. 
\begin{itemize}
    \item \textbf{Offices:} Featuring narrow corridors, cubicles, and meeting rooms, challenging the agent's fine-grained motion control.
    \item \textbf{Shopping Malls:} Spanning vast areas (\textbf{3.02 $\bm{km^2}$}) with complex topology, multiple floor levels, and reflective surfaces (e.g., glass storefronts).
    \item \textbf{Transit Stations:} Characterized by channelized pedestrian traffic, navigational bottlenecks (e.g., turnstiles), and large-scale waiting halls interspersed with narrow transit corridors.
\end{itemize}

\subsubsection{Outdoor Environments: Static Scans and Dynamic City}
The outdoor domain introduces unique challenges such as traffic rules, unstructured terrain, and vast spatial scales.

\paragraph{Real-World Scans (Intersection, Park)}
We utilize high-precision LiDAR and photogrammetry to reconstruct \textbf{37} real-world outdoor scenes.
\begin{itemize}
    \item \textbf{Intersections:} These scenes allow the agent to learn the geometry of crosswalks, vehicle lanes, and sidewalks. Our road network annotation strictly forbids the agent from entering vehicle lanes, enforcing traffic rule compliance.
    \item \textbf{Parks:} These unstructured park environments (\textbf{8 scenes}) cover narrow walking paths and open plazas.
\end{itemize}

\paragraph{Dynamic Virtual City (SocCity)}
To complement static scans, we employ SocCity \cite{chen2025socialnav}, a large-scale (\textbf{3.37 $\bm{km^2}$}) virtual urban environment. Unlike static 3DGS scans, SocCity supports dynamic simulation of vehicles and pedestrians. The occupancy map here is hierarchically provided to distinguish between pedestrian walkways, crosswalks, vehicle roads and dynamic obstacles, enabling the training of \textit{Socially-Aware Flow Exploration} (SAFE-GRPO) \cite{chen2025socialnav}.

\subsection{\MODEL Trajectories Dataset}
\label{sec:trajectory_data}

Leveraging the unified navigation graphs and high-fidelity environments established in Sec.~\ref{sec:data_source}, we construct the \textbf{\MODEL Trajectory Dataset}, a massive, standardized corpus of sensorimotor demonstrations comprising approximately \textbf{16.9 million} training trajectories across five navigation paradigms. Unlike prior works that rely on fragmented datasets with task-specific formats, this collection unifies Point-Goal, Instruction-Following, Object-Goal, POI-Goal, and Person-Following tasks into a shared action representation. By deploying optimal planners on navigation graphs and aggregating large-scale human/robot video logs, we align diverse high-level intent signals with precise, continuous trajectory distributions, providing the essential supervision for the Action Expert to learn a generalized motion policy.

\begin{figure*}[t]
    \centering
    \includegraphics[width=0.95\linewidth]{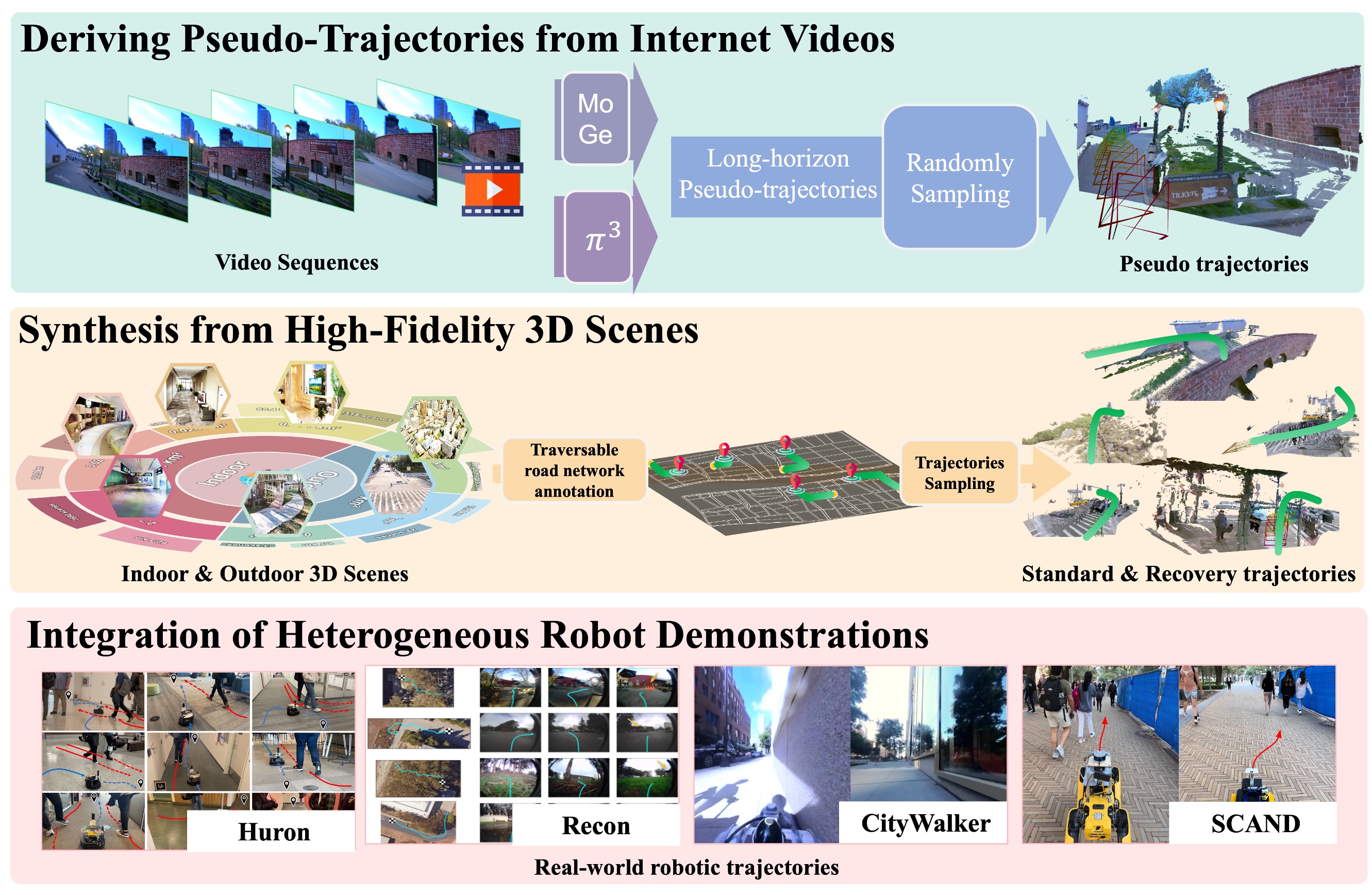}
    \caption{\textbf{Point-Goal Navigation Data Pipeline.} Our \textbf{4.0M} point-goal trajectory dataset aggregates three complementary streams: (Top) \textbf{2.0M} pseudo-trajectories from first-person internet videos via 3D structure recovery and metric alignment; (Middle) \textbf{1.7M} synthetic trajectories from high-fidelity 3D scenes with annotated navigation graphs; (Bottom) \textbf{340K} real-world robot demonstrations from heterogeneous datasets, providing ground-truth physical dynamics.}
    \label{fig:pointgoal}
\end{figure*}

\subsubsection{Point-Goal Trajectories}

Point-goal navigation serves as the foundational locomotion primitive in our data engine, focusing on the agent's ability to reach precise metric coordinates while maintaining environmental and social awareness. To cultivate a robust and generalizable navigation foundation, we curate a large-scale collection of approximately \textbf{4.0 million} point-goal navigation trajectories. This dataset is synthesized through a multi-source data engine that transforms raw 3D scenes, uncurated internet videos, and heterogeneous robotic logs into standardized navigation episodes (See Fig.\ref{fig:pointgoal}). The construction process focuses on ensuring visual diversity, metric accuracy, and kinematically feasible motion priors.

\paragraph{Deriving Pseudo-Trajectories from Internet Videos} 
To capture the long-tail distribution of global architectural styles and weather conditions, we process \textbf{2.0 million} pseudo-trajectories from a curated collection of first-person urban exploration videos. We employ a scalable vision-to-trajectory pipeline: 
(1) \textbf{Structure Recovery}: Utilizing $\pi^{3}$~\cite{wang2025pi} for dense 3D reconstruction from monocular streams; 
(2) \textbf{Metric Alignment}: Applying MoGe~\cite{wang2025moge} to resolve scale ambiguity and align the reconstructed paths with real-world metric units; 
(3) \textbf{Episode Synthesis}: Sampling diverse point-goal pairs along the camera's motion manifold to generate kinematically consistent navigation samples. This approach allows the data engine to ingest ``passive'' observation data and convert it into ``active'' navigation knowledge.

\paragraph{Synthesis from High-Fidelity 3D Scenes} 
We leverage a vast repository of 3D scenes in our collection and the large-scale dynamic SocCity \cite{chen2025socialnav}. Point-goal episodes are generated by sampling reachable coordinate pairs $(s, g)$ and computing optimal paths using path-finding algorithms. Crucially, we incorporate recovery trajectories by initializing agents in off-path or near-collision states, forcing the model to learn error-correction behaviors essential for real-world robustness. Finally, \textbf{1.7 million} training samples are generated through these 3D scene assets.

\paragraph{Integration of Heterogeneous Robot Demonstrations} 
We incorporate \textbf{340K} high-quality trajectories from diverse real-world robotic platforms (e.g., SCAND~\cite{karnan2022socially}, HuRoN \cite{hirose2023sacson}, Recon~\cite{shah2022rapid}, and CityWalker \cite{liu2025citywalker} teleoperation data). Unlike simulated or video-derived data, these trajectories provide ground-truth physical dynamics and sensor noise characteristics. We standardize these multi-modal logs into a unified point-goal format, ensuring that the model internalizes the intricate constraints of real-world hardware, such as non-holonomic motion limits and sensor-to-actuation latency. 

By aggregating these three streams, the data engine provides a comprehensive trajectory distribution that spans from idealistic optimal paths to complex, noise-perturbed real-world maneuvers, forming the backbone of our model's navigation intelligence. More details of this pipeline is reported in our SocialNav~\cite{chen2025socialnav}.

\subsubsection{Instruction-Following Trajectories}

Within the Visual Language Navigation (VLN) framework \cite{zhang2025embodiednavigationfoundationmodel, wei2025streamvln, cheng2024navila,zhang2024uni, zhang2024navid, long2024instructnav}, instruction following is defined as a task setting where an agent must navigate a 3D environment by continuously processing egocentric visual observations to reach a destination specified by natural language instructions. This paradigm emphasizes the alignment and execution of the navigation action derived from complex linguistic input. Its significance extends beyond simply reaching the target; it critically evaluates the agent's adherence to the specific paths and procedural milestones described in the instruction, providing a more authentic assessment of language grounding and instruction-following capabilities. To enable the model to comprehend natural language and plan the corresponding trajectories, we constructed the following datasets.

\paragraph{VLN-CE R2R/RxR} This dataset is built within the VLN-CE (Continuous Environment) framework \cite{krantz_vlnce_2020} using Matterport3D \cite{chang2017matterport3d} scenes. For each training instance, we recorded the navigation trajectory waypoints, the corresponding natural language instruction, and panoramic observations comprising three rendered views (front, left, and right) at each step.

The VLN-CE R2R \cite{krantz_vlnce_2020} component consists of approximately 10K continuous clips, each representing a shortest-path navigation task paired with an instruction. Employing a teacher forcing protocol, we unrolled these trajectories into individual time-step samples and applied data balancing to normalize the action distribution. This process filtered the initial pool of 600K samples down to 200K retained training trajectory samples.

VLN-CE RxR \cite{ku2020room} represents a significant expansion over R2R in both scale and complexity. Compared to R2R, RxR features longer path lengths, more complex topological structures, and instructions characterized by higher linguistic density—incorporating frequent and semantically diverse references to landmarks and complicated spatial relationships. Consequently, it offers richer supervision for long-horizon language grounding and continuous decision-making. This subset contains approximately 20K continuous clips. Following a similar balancing protocol, we curated 1.3 million training samples from an initial pool of 1.8 million. In total, the combined VLN-CE R2R/RxR dataset comprises \textbf{1.5 million} training trajectory samples.

\begin{figure*}[t]
    \centering
    \includegraphics[width=\linewidth]{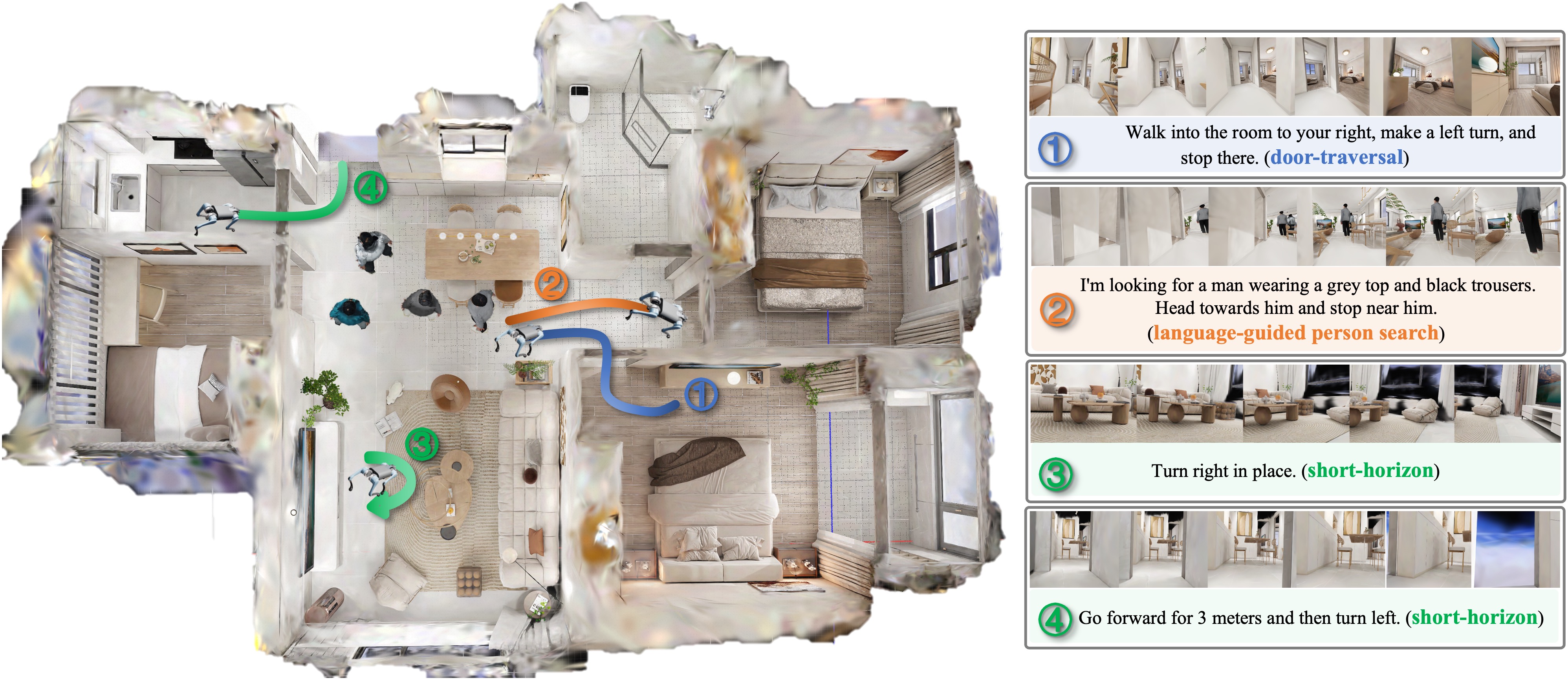}
    \caption{\textbf{Diverse Instruction-Following Tasks in InteriorGS \cite{InteriorGS2025}.} We illustrate three representative navigation scenarios designed to enhance instruction-following capabilities: (a) Door-Traversal for navigating narrow passages, (b) Language-Guided Person Search, and (c) Short-Horizon navigation for atomic movement primitives.}
    \label{fig:insturction_subtasks}
\end{figure*}

\paragraph{Door-Traversal} To address the specific challenges associated with navigating narrow passages and bottlenecks, we synthesized 6,000 door-traversal clips, which were subsequently processed into \textbf{0.3 million} training trajectory samples. Leveraging InteriorGS \cite{InteriorGS2025}, we sampled initial poses in the vicinity of doorways while enforcing a diverse distribution of post-traversal orientations and travel distances. Additionally, we annotated each trajectory with multiple distinct instructions to maximize diversity in both geometric configurations and linguistic variability. A schematic illustration of the data acquisition process is presented in Fig.~\ref{fig:insturction_subtasks}.

\paragraph{Language-Guided Person Search} We generated a diverse library of digital human assets, consisting of 200 3DGS avatars for the InteriorGS \cite{InteriorGS2025} environment and 2,000 polygonal mesh characters for the Habitat simulator \cite{savva2019habitat}. Descriptions for these avatars were synthesized from attribute combinations (e.g., clothing, gender, pose) and subsequently refined using LLMs to ensure natural linguistic flow. For each training episode, we populated the scene with 2 to 5 randomly instantiated avatars, designating one as the navigation target based on a specific language query. We then initialized the agent at a random starting pose and computed ground-truth trajectories using the geodesic shortest path to the target character (Fig.~\ref{fig:insturction_subtasks}). In total, this process yielded 4,000 training clips, culminating in a dataset of \textbf{0.2 million} trajectory samples.

\paragraph{Short-Horizon} Focusing on atomic navigation primitives characterized by short horizons and explicit metric goals, we curated 12,000 clips within InteriorGS \cite{InteriorGS2025}, which were subsequently processed into \textbf{0.8 million} training trajectory samples. This dataset encompasses rotational actions (e.g., turn left, turn around, turn left 60 degrees), translational movements (e.g., move forward 1 meter, move backward 3 meters), and compound sequences that integrate both rotation and translation as shown in Fig.~\ref{fig:insturction_subtasks}. These samples are specifically designed to enhance the model's fine-grained execution capabilities. Furthermore, the inclusion of these fundamental short-horizon tasks serves to effectively regularize and stabilize the model training process.

\subsubsection{Object-Goal Trajectories}

Object-Goal Navigation (ObjectNav) requires an agent to actively search for and navigate to an instance of a specific object category within an unseen environment. Unlike traditional Point-Goal Navigation (PointNav) tasks, ObjectNav relies heavily on semantic reasoning and the effective integration of multimodal information. Furthermore, the heterogeneity of indoor layouts and the complexity caused by environmental occlusions exacerbate the difficulty of the task, posing significant challenges to the model's generalization and robustness in diverse real-world scenarios.

\paragraph{HM3D and OVON} For the HM3D ObjectNav task \cite{yokoyama2024hm3d}, we focused on 6 target object categories across 80 scenes, constructing a training set of 0.4 million trajectories derived from 10K expert clips. To further strengthen the model's generalization capabilities, we incorporated the Open-Vocabulary ObjectNav (OVON) benchmark \cite{yokoyama2024hm3d}. Unlike standard ObjectNav, which operates on a closed set of categories, OVON requires the agent to locate targets conditioned on free-form language queries. This task rigorously assesses performance across unseen categories, synonyms, and semantically distant objects. Spanning 145 scenes, we sampled 10 starting poses for each object instance, generating a comprehensive navigation dataset comprising 35K episodes and a total of 1.4 million training trajectories. In total, we obtained \textbf{1.8 million} trajectories.

\paragraph{OVON-sub} To mitigate the complexity associated with long-horizon planning and object occlusion in standard OVON tasks, we constructed a short-range navigation subset specifically designed to enhance the model's visual-semantic alignment. We generated this subset by processing the original ground-truth trajectories with a specific truncation strategy: we identify the onset of target visibility—the first timestep where the target object enters the agent's Field of View (FoV); and retain the trajectory segment from this moment until the episode's termination. This process yielded 6,000 training clips including \textbf{0.2 million} training trajectory samples.

\begin{figure*}[t]
  \centering
  \begin{minipage}{0.58\textwidth}
    \centering
    \includegraphics[width=\linewidth]{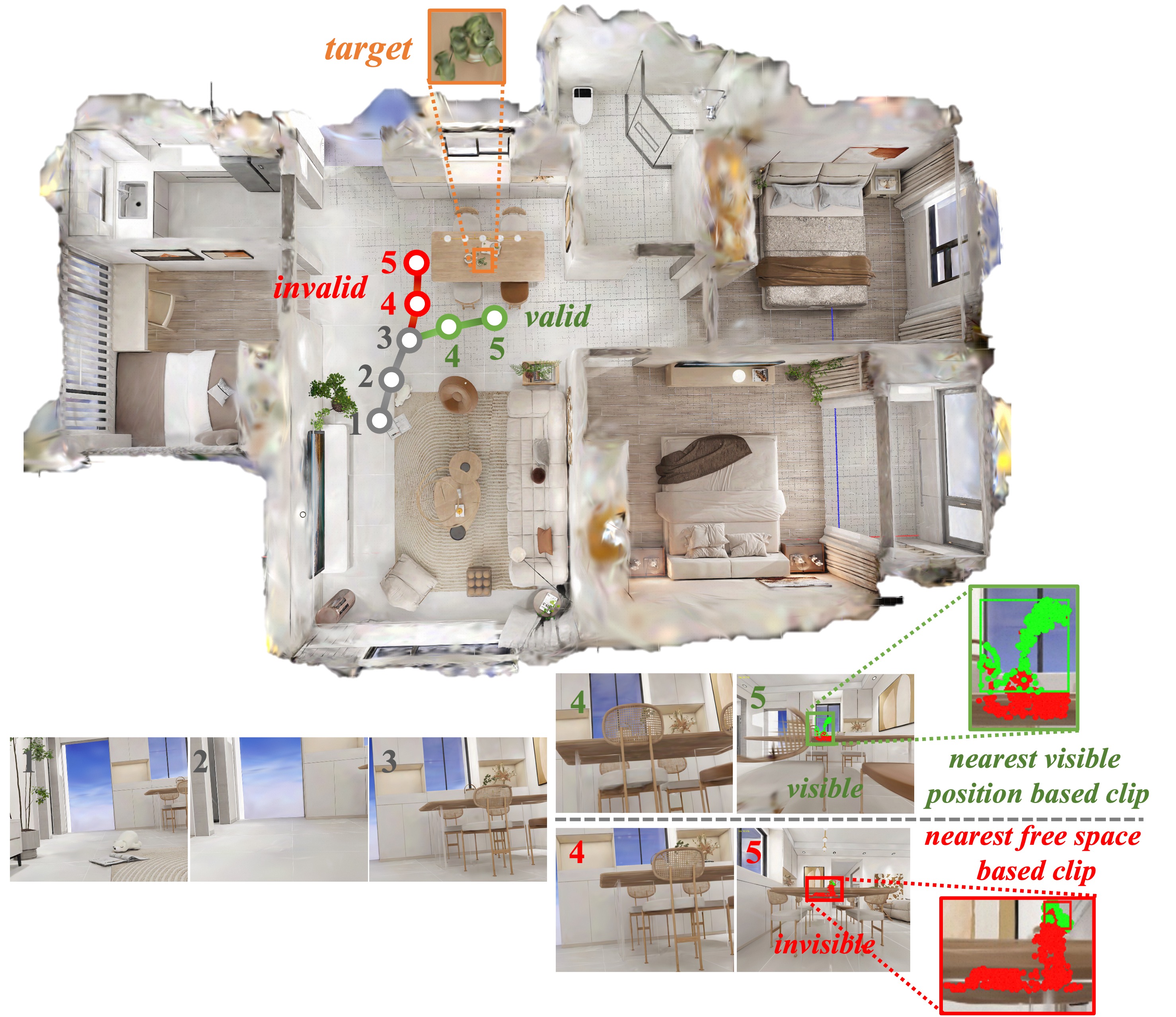}
    \caption{\textbf{Trajectory generation pipeline for Object Goal tasks in InteriorGS \cite{InteriorGS2025}.} Our method selects the nearest visible position as the clip endpoint, ensuring cleaner training data compared to the naive approach of using the nearest free space.}
    \label{fig:interiorgs_gt}
  \end{minipage}
  \hfill
  \begin{minipage}{0.40\textwidth}
    \centering
    \includegraphics[width=\linewidth]{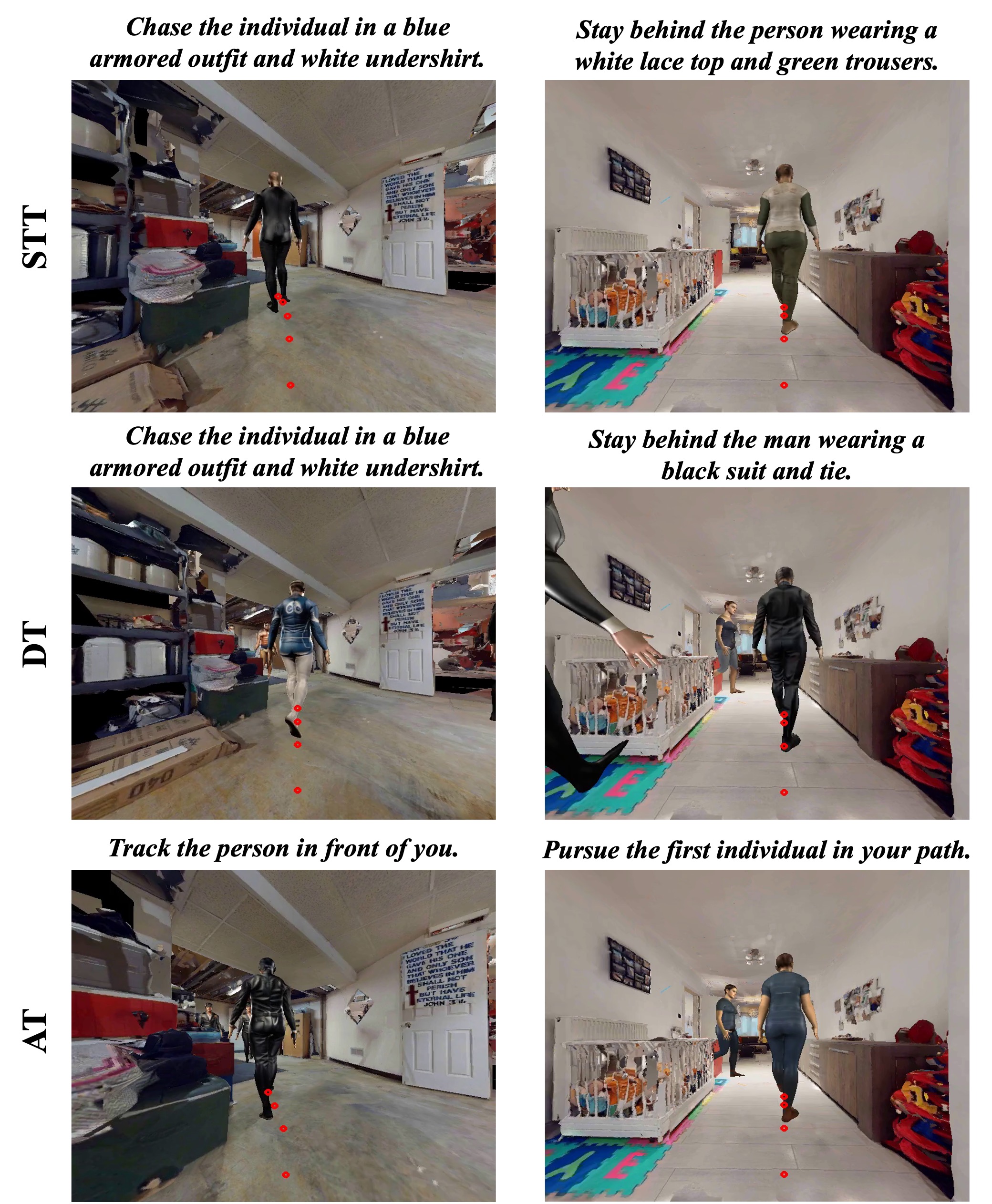}
    \caption{\textbf{Trajectory generation pipeline for Person-Following tasks.} We track human trajectories and sample viewpoints along the path to create training data for person-following navigation.}
    \label{fig:tracking}
  \end{minipage}
\end{figure*}

\paragraph{InteriorGS} Compared to HM3D \cite{ramakrishnan2021habitat}, InteriorGS \cite{InteriorGS2025} provides superior photorealistic rendering fidelity, featuring 1,000 scenes annotated with bounding boxes for over 700 object categories. However, the absence of pre-defined visible position annotations precludes the direct application of the standard HM3D ObjectNav data generation protocol. To bridge this gap, we developed a custom trajectory generation pipeline driven by object visibility estimation as shown in Figure~\ref{fig:interiorgs_gt}. 
Finally, we construct a training set of 40K clips with \textbf{1.6 million} trajectory samples.

\subsubsection{POI-Goal Trajectories}
We have proposed the POI-Goal task along with its corresponding dataset; detailed definitions of the task and specifics of dataset construction can be found in~\citet{zhao2026bridgingindooroutdoorgapvisioncentric}. This task aims to bridge outdoor and indoor environments, enabling embodied agents to navigate from open, coarse-grained outdoor scenes into structured, fine-grained indoor spaces.

Given a single input image, we first employ image segmentation and depth estimation techniques to generate a local occupancy grid map. We then use the A* algorithm to plan a trajectory from the starting position to the POI, which serves as the ground-truth path. Next, we feed both the input image and the generated trajectory into the Wan2.1-I2V Diffusion Transformer video generation model \cite{wan2025wan} to synthesize first-person-view observation videos that simulate the navigation process of an embodied agent. Finally, \textbf{2.5 million} training trajectories are sampled from these generated videos.

\subsubsection{Person-Following Trajectories}

We referenced the dataset creation method of TrackVLA \cite{wang2025trackvla}. However, since its training data was not fully open sourced, we generated our own training data based on its open-sourced humanoid avatar motion trajectories (See Fig.\ref{fig:tracking}). 
We generated synthetic tracking sequences through three distinct proximity configurations (2.0m, 1.5m, and 1.2m target distances) to emulate real-world tracking scenarios. For each distance configuration, we employed the A* algorithm to compute optimal waypoints while maintaining stable tracking observations. Following TrackVLA's established taxonomy, we generated three tracking challenge categories (Single-Target Tracking/STT, Distracted Tracking/DT, Ambiguity Tracking/AT) for each distance configuration. Each category-distance combination contains 400K samples, yielding 3.6 million base training instances (3 distances × 3 categories × 400K). To address target absence scenarios where no tracked subject appears in the field of view, we supplemented the dataset with 400K target-absent cases. The final composite dataset comprises \textbf{4.0 million} training samples, with each sample containing current frame images, historical frame sequences, their corresponding future trajectories, and associated navigation commands.

\subsection{\MODEL Reasoning Dataset}
\label{sec:reasoning_dataset}

\begin{figure}[htbp]
    \centering
    \includegraphics[width=\linewidth]{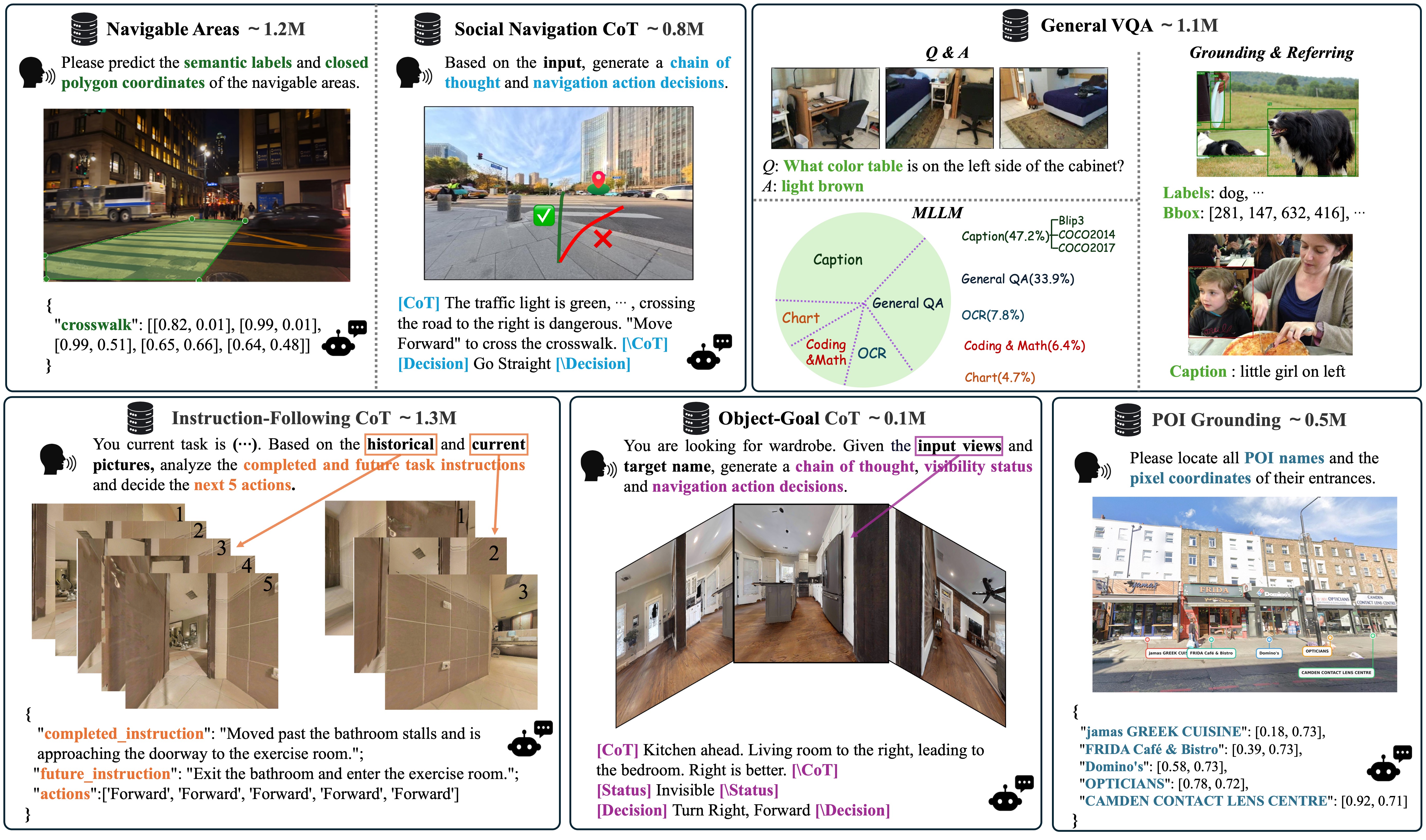}
    \caption{\textbf{Examples from the \MODEL Reasoning Dataset.} The dataset encompasses diverse reasoning tasks including Navigable Areas Analysis, Social Navigation CoT, Instruction-Following Reasoning, Object-Goal Reasoning, POI Grounding, and General VQA.}
    \label{fig:reasoning}
\end{figure}

While the Trajectory Dataset provides the sensorimotor experience (how to move), it lacks the explicit causal logic required for robust decision-making in unseen environments. To bridge this gap, we introduce the \textbf{\MODEL Reasoning Dataset}, a large-scale corpus comprising \textbf{5.0 million} samples designed to activate the high-level cognitive capabilities of the VLA Brain. It compels the model to explicitly verbalize its understanding of the scene geometry, semantic landmarks, and social norms when committing to an action. The dataset is structured hierarchically, progressing from general navigation skills to task-specific reasoning, and general world knowledge.

\subsubsection{General Navigation Skills}
\label{sec:general_skills}

This subset instills the fundamental spatial and social common sense required for any mobile agent, regardless of its specific mission. We adopt the data generation pipeline proposed in our SocialNav~\cite{chen2025socialnav}.

\paragraph{Navigable Areas Analysis (1.2M Samples)}
A core competency for any embodied agent is distinguishing between ``physically traversable'' and ``socially traversable'' areas. To ground the agent's visual perception in physical affordances, we construct a dataset comprising \textbf{1.2 million} ego-centric images collected from diverse outdoor video streams. For each image, we manually annotate precise polygons that delineate socially compliant zones (e.g., sidewalks, crosswalks) while strictly excluding non-navigable areas (e.g., roadways, lawns). The model is trained to output the semantic labels and polygon coordinates of these regions given a visual observation. This task helps ensure that the generated waypoints in downstream tasks always reside within safe boundaries, effectively constraining the action space with social norms.

\paragraph{Social Navigation Chain-of-Thought (0.8M Samples)}
To enable the model to handle complex social interactions (e.g., yielding to pedestrians, obeying traffic lights), we construct a massive dataset of navigation rationales. Leveraging the automated pipeline from SocialNav \cite{chen2025socialnav}, we utilize a powerful VLM (Qwen-VL-Max \cite{Qwen-VL}) as a teacher agent to generate structured Chain-of-Thought (CoT) rationales. Each sample contains the reasoning process that explicitly explains the logic behind a decision (e.g., \textit{``The traffic light is red, so I must wait''}).

\subsubsection{Instruction-Following Reasoning}

In long-horizon navigation, embodied agents often struggle to align visual observations with instruction progress, leading to trajectory drift and task failure. Leveraging the sequential structure of global instructions, we decompose long-horizon tasks into atomic sub-instructions. For each sub-instruction, we identify its spatiotemporal termination point, defined as a ``milestone node''. These nodes form a Visual CoT that explicitly grounds the execution process. We automate this pipeline using MLLMs, specifically Gemini-3 Pro \cite{google_gemini3pro_models_docs} and Qwen3-VL \cite{yang2025qwen3}. By processing 30,000 clips from the VLN-CE R2R/RxR datasets \cite{krantz_vlnce_2020, ku2020room}, we generated \textbf{1.3 million} training samples. Each sample consists of current and historical observations, completed instruction, future instruction, and predicted action. More pipeline details are in our NavForesee~\cite{liu2025navforesee}.

\subsubsection{Object-Goal Reasoning}

End-to-end policies, typically trained with sparse action supervision, often overfit to implicit biases within the training distribution. They lack explicit constraints on intermediate reasoning steps, such as target visibility or spatial relationships. This limitation severely hampers generalization across open-vocabulary targets and unseen scenarios. To address this, we utilized the OVON dataset \cite{yokoyama2024hm3d} to construct Object Reasoning, a large-scale CoT dataset.

Specifically, we prompt MLLMs to generate structured reasoning chains based on panoramic observations and target information. These chains encompass four key components: (1) target visibility assessment, (2) spatial relationship analysis, (3) path planning, and (4) the corresponding atomic action sequence. To ensure data quality, we implemented a consistency check mechanism. We filter out hallucinations by verifying that the generated action sequences align with ground-truth trajectories. Ultimately, we curated \textbf{0.1 million} high-quality samples. This design provides interpretable reasoning for decision-making and effectively bridges the gap between high-level semantic intent and low-level control.

\subsubsection{POI Grounding}
\label{sec:poi_grounding}

Standard object detection datasets often lack the precision required for navigation, specifically in linking semantic Point-of-Interest (POI) goals to physical entry points. To bridge this gap, we introduce the POI Grounding Dataset. 

We curated a subset of street-view images from the BridgeNav Dataset~\cite{zhao2026bridgingindooroutdoorgapvisioncentric}, covering diverse urban scenes. Unlike traditional facade-level labeling, we focus on precise entry point localization. Utilizing Qwen3-VL-Plus~\cite{yang2025qwen3}, we automatically generate structured annotations as tuples: $\langle \text{POI Name}, \text{Entrance Coordinates } (u, v) \rangle$. This process yields \textbf{0.5 million} Visual Question Answering (VQA) pairs, providing the spatial grounding necessary for the \MODEL to translate semantic POI targets into executable waypoints.

\subsubsection{General VQA}
To preserve general visual-linguistic representations and enhance generalization in unseen environments, we incorporated a diverse set of VQA and Embodied Question Answering (EQA) datasets. Specifically, we utilized Blip3 \cite{chen2025blip3o}, COCO2014 \cite{lin2014microsoft}, COCO2017 \cite{lin2015microsoftcococommonobjects} and MAmmoTH-VL~\cite{guo2025mammoth} to provide general commonsense knowledge. To bolster the model's grounding and referring capabilities, we included the RefCOCO series~\cite{kazemzadeh2014referitgame} and Objects365~\cite{shao2019objects365}. Furthermore, we integrated R2R-EnvDrop~\cite{tan2019learning} and ScanQA~\cite{azuma2022scanqa}. These indoor embodied datasets are critical for developing spatial awareness and trajectory planning skills. Collectively, we curated a total of \textbf{1.1 million} training samples from these sources.
\section{Training Recipe}
\label{sec:training_recipe}

Training a unified foundation model across five distinct navigation tasks presents a significant challenge: \textit{multimodal alignment}. The model must balance high-level semantic reasoning (e.g., understanding the relationship between observed images and navigation instructions) with low-level waypoints-based control.

To address this, we propose a three-stage curriculum learning pipeline: \textbf{Cognitive Warm-up}, \textbf{Unified Sensorimotor SFT}, and \textbf{Post-Training Value Alignment via SAFE-GRPO}. This progressive approach ensures that \MODEL first acquires a robust understanding of the world before learning to act within it.

\subsection{Phase 1: Cognitive Warm-up}

Before learning strictly ``how to move'', the agent must learn ``what to see'' and ``how to reason.'' In this phase, we utilize the \textbf{\MODEL Reasoning Dataset} to align the LLM backbone with the embodied domain. By freezing the Vision Encoder and text tokenizer, we fine-tune the LLM core using Next Token Prediction (NTP) loss across diverse tasks (e.g., general navigation skills, specific navigation reasoning tasks and general VQAs). This process enables the model to interpret visual scenes, ground textual instructions to physical entities, and perform complex spatial reasoning. Throughout this phase, the Action Expert remains frozen, ensuring that gradients are dedicated exclusively to optimizing the visual-linguistic representations and cognitive foundations of the model.

\subsection{Phase 2: Unified Sensorimotor SFT}

Once the cognitive backbone is established, we introduce the \textbf{\MODEL Trajectory Dataset}. This phase unifies all five navigation tasks into a single multi-task training regime. To preserve the reasoning capabilities acquired in the previous phase and prevent catastrophic forgetting, we employ a mixed-training strategy by augmenting the trajectory data with a replay buffer of reasoning data at an approximate ratio of 20\%. During this phase, we perform \textbf{Dual-Head Optimization} to jointly optimize the AR Head (autoregressive reasoning) and the Action Expert (flow matching). The training objective is defined by the joint loss function:
\begin{equation}
    \mathcal{L}_{Phase2} = \lambda_{txt} \mathcal{L}_{NTP}(\theta_{brain}) + \lambda_{flow} \mathcal{L}_{CFM}(\theta_{action} | \theta_{brain})
\end{equation}
where $\mathcal{L}_{NTP}$ denotes the cross-entropy loss for text generation and $\mathcal{L}_{CFM}$ represents the Conditional Flow Matching loss for trajectory generation. This integrated approach grounds high-level semantic plans into precise  continuous physical actions, resulting in a unified policy capable of executing diverse and complex embodied tasks.

\subsection{Phase 3: Post-Training Value Alignment via SAFE-GRPO}

Although Phase 2 equips the model with general navigation capabilities via Imitation Learning (IL), IL inherently suffers from a limitation: it captures the surface-level statistics of expert behaviors but fails to master the causal structure underlying normative conduct in complex social environments. For instance, an agent might learn to walk on a sidewalk, but fail to understand why deviating into a flowerbed or a vehicle lane is strictly prohibited.

To bridge this gap, we employ \textbf{SAFE-GRPO} (Socially-Aware Flow Exploration GRPO) \cite{chen2025socialnav}, a flow-based reinforcement learning framework designed to explicitly enforce social compliance.

During this phase, we freeze the Brain model to preserve the semantic understanding acquired in previous stages, and exclusively fine-tune the Action Expert. The training is conducted with the expert trajectories from SocCity \cite{chen2025socialnav}, which provides the rich annotations necessary to calculate precise rewards.

The policy is optimized to maximize a composite reward function $\mathcal{R}$ that balances social compliance with navigation efficiency:

\begin{equation}
    \mathcal{R} = w_{soc} \mathcal{R}_{social} + w_{exp} \mathcal{R}_{expert} + w_{sm} \mathcal{R}_{smooth} + w_{eff} \mathcal{R}_{eff}
\end{equation}

\begin{itemize}
    \item \textbf{Social Compliance Reward ($\mathcal{R}_{social}$):} This is the primary signal for the alignment of values. It is derived from a ground-truth \textit{semantic occupancy map} provided by the SocCity environment. The agent incurs a heavy penalty if its predicted trajectory traverses non-navigable or socially restricted zones (e.g., lawns, restricted lanes, walking humans), forcing the model to internalize the causal rule: ``navigable geometry $\neq$ socially acceptable path''.
    
    \item \textbf{Expert Similarity Reward ($\mathcal{R}_{expert}$):} To prevent reward hacking, we maintain a regularization term that encourages the policy to stay within a reasonable distribution of the expert demonstrations.
    
    \item \textbf{Smoothness \& Efficiency ($\mathcal{R}_{smooth}, \mathcal{R}_{eff}$):} These terms penalize jerky movements and encourage progress toward the goal, ensuring the generated trajectories remain feasible for real-world robot execution.
\end{itemize}

Through this value-aligned post-training, \MODEL ensures strict adherence to social norms when executing all navigation tasks.
\section{Evaluation}

To empirically validate \MODEL as a generalist foundation model, we conduct a comprehensive evaluation across five distinct navigation paradigms. By benchmarking against task-specific state-of-the-art methods on standard platforms, we demonstrate that \MODEL not only unifies these disparate tasks within a single architecture but also achieves superior performance through its reasoning-driven policy.

\subsection{Point-Goal Task}
\label{sec:point_goal_eval}

\begin{table*}[t]
    \centering
    
    \caption{\textbf{Open-Loop Point-Goal Evaluation on CityWalker Benchmark~\cite{liu2025citywalker}.} We evaluate MAOE metric in each critical scenario for all methods. Percentages under scenarios indicate their data proportions. The ``Mean'' column shows scenario means averaged over six scenarios; ``All'' shows sample means over all data samples.}
    \label{tab:pointopen}
    \begin{tabular}{l|cccccccc}
    \toprule

        \multirow{2}{*}{\textbf{Method}} & \multirow{2}{*}{\textbf{Mean}} & \textbf{Turn} & \textbf{Crossing} & \textbf{Detour} & \textbf{Proximity} & \textbf{Crowd} & \textbf{Other} & \textbf{All} \\ 

        & & 8\% & 12\% & 12\% & 6\% & 7\% & 55\% & 100\% \\ 
        \midrule
        GNM~\cite{shah2023gnm}
         & 16.2 & 31.1 & 14.8 & \underline{12.5} & 14.7 & 12.8 & 11.0 & 12.1\\

        ViNT~\cite{shah2023vint}
         & 16.5 & 31.1 & 15.4 & 12.9 & 14.8 & 13.3 & 11.6 & 12.6\\

        NoMaD~\cite{sridhar2024nomad}
         & 19.1 & 35.1 & 18.5 & 15.6 & 18.1 & 14.3 & 12.8 & 12.1\\

        CityWalker~\cite{liu2025citywalker} 
         &  \underline{15.2} &  \underline{26.6} &  \underline{14.1}  &  13.9 &  \underline{14.3} &  \underline{12.0} &  \underline{10.4} &  \underline{11.5}\\ \midrule
        \textbf{\MODEL}
         &  \textbf{11.2} & \textbf{21.3} &  \textbf{9.8}  &  \textbf{12.8} &  \textbf{8.1} &  \textbf{8.8} &  \textbf{6.3} &  \textbf{7.6}\\ 
    \bottomrule
    \end{tabular}
\end{table*}

\begin{table*}[t]
\centering

\caption{\textbf{Point-Goal Performance Comparison on the Closed-Loop SocNav Benchmark \cite{chen2025socialnav}}. \MODEL demonstrates superior navigation performance and social compliance across all metrics.}
\label{tab:pointclose}

{

\newcolumntype{C}{>{\centering\arraybackslash}p{1.5cm}}

\begin{tabular}{l|CCC|CC} 
    \toprule
    \multirow{2}{*}{\textbf{Method}} & \multicolumn{3}{c|}{\textbf{Navigation Performance}} & \multicolumn{2}{c}{\textbf{Social Compliance}} \\
     & \textbf{SR$\uparrow$} & \textbf{RC$\uparrow$} & \textbf{SPL$\uparrow$} & \textbf{DCR$\uparrow$} & \textbf{TCR$\uparrow$}  \\
    \midrule
    GNM*~\cite{shah2023gnm} & 43.3 & 62.4 & 37.0 & 26.5 & 28.7 \\
    ViNT*~\cite{shah2023vint} & 45.6 & \underline{66.2} & 39.5 & 31.4 & 33.8   \\
    NoMaD*~\cite{sridhar2024nomad} & 41.1 & 60.5 & 35.4 & 29.5 & 31.6  \\
    CityWalker~\cite{liu2025citywalker} & \underline{47.8} & 64.7 & \underline{44.7} & \underline{36.1} & \underline{36.6} \\ \midrule
   \textbf{\MODEL} & \textbf{88.3}  & \textbf{92.1} & \textbf{79.2} & \textbf{85.1} & \textbf{85.4}  \\
    \bottomrule
\end{tabular}
}
\end{table*}

Following the rigorous evaluation protocol established in SocialNav~\cite{chen2025socialnav}, we assess the Point-Goal navigation performance of \MODEL through a comprehensive combination of open-loop benchmarks and closed-loop simulations.

\paragraph{Evaluation Setup and Metrics}
We conduct the open-loop evaluation on the \textbf{CityWalker Benchmark}~\cite{liu2025citywalker}. The primary metric is Maximum Average Orientation Error (MAOE), which measures the maximum angular deviation between the predicted and ground-truth actions over the future horizon, serving as a proxy for human-likeness.
For closed-loop evaluation, we utilize the \textbf{SocNav Benchmark}~\cite{chen2025socialnav}, deploying the agent in the unseen and high-fidelity environments. We report standard navigation metrics: Success Rate (SR), Route Completion (RC), and Success weighted by Path Length (SPL). Crucially, to quantify social awareness, we employ Distance Compliance Rate (DCR) and Time Compliance Rate (TCR), which calculate the percentage of distance and time the agent spends strictly within socially traversable zones (e.g., sidewalks) versus restricted areas (e.g., lawns or roadways).

\paragraph{Results and Analysis}
The quantitative results are presented in Table~\ref{tab:pointopen} and Table~\ref{tab:pointclose}.
In the open-loop setting, \MODEL achieves a mean MAOE of \textbf{11.2}, significantly outperforming the previous state-of-the-art CityWalker (15.2) and other baselines.
In the closed-loop setting, \MODEL demonstrates remarkable robustness and social adherence. It attains an \textbf{88.3\%} Success Rate, nearly doubling the performance of the baseline (47.8\%). More importantly, in terms of social compliance, \MODEL achieves a DCR of \textbf{85.1\%} compared to 36.1\% for the baseline. This confirms that our model does not merely reach the goal but does so by actively respecting social norms, verifying the effectiveness of our hierarchical ``Brain-Action'' architecture and the SAFE-GRPO alignment strategy.

\subsection{Instruction-Following Task}

\begin{table*}[t]
    \centering
    \caption{\textbf{Instruction-following evaluation on VLN-CE benchmarks \cite{krantz_vlnce_2020, ku2020room}.} We compare \MODEL with state-of-the-art methods on R2R-CE and RxR-CE Val-Unseen splits. Methods marked with * use the waypoint predictor from \citet{hong2022bridging}. \MODEL achieves superior performance using panoramic observations, demonstrating strong generalization in continuous navigation environments.}
    \label{tab:r2r and rxr}
    
    \begin{adjustbox}{max width=\textwidth}
    \begin{tabular}{l|cccc|cccc|ccc}
    \toprule
    \multirow{2}{*}{\textbf{Method}} & \multicolumn{4}{c|}{\textbf{Observation}} & \multicolumn{4}{c|}{\textbf{R2R-CE Val-Unseen}} & \multicolumn{3}{c}{\textbf{RxR-CE Val-Unseen}} \\
    
      & \textbf{S.RGB} & \textbf{Pano.} & \textbf{Depth} & \textbf{Odo.} & \textbf{NE$\downarrow$} & \textbf{OS$\uparrow$} & \textbf{SR$\uparrow$} & \textbf{SPL$\uparrow$} & \textbf{NE$\downarrow$} & \textbf{SR$\uparrow$} & \textbf{SPL$\uparrow$} \\
    
    \midrule
    HPN+DN*~\citep{krantz2021waypoint} & & \checkmark & \checkmark & \checkmark & 6.31 & 40.0 & 36.0 & 34.0 & - & - & - \\
    CMA*~\citep{hong2022bridging} & & \checkmark & \checkmark & \checkmark & 6.20 & 52.0 & 41.0 & 36.0 & 8.76 & 26.5 & 22.1 \\
    Sim2Sim*~\citep{krantz2022sim} & & \checkmark & \checkmark & \checkmark & 6.07 & 52.0 & 43.0 & 36.0 & - & - & - \\
    GridMM*~\citep{wang2023gridmm} & & \checkmark & \checkmark & \checkmark& 5.11 & 61.0 & 49.0 & 41.0 & - & - & - \\
    DreamWalker*~\citep{wang2023dreamwalker} & & \checkmark & \checkmark & \checkmark & 5.53 & 59.0 & 49.0 & 44.0 & - & - & - \\
    Reborn*~\citep{an20221st} & & \checkmark & \checkmark & \checkmark & 5.40 & 57.0 & 50.0 & 46.0 & 5.98 & 48.6 & 42.0 \\
    ETPNav*~\citep{an2024etpnav} & & \checkmark & \checkmark & \checkmark & 4.71 & 65.0 & 57.0 & 49.0 & 5.64 & 54.7 & 44.8 \\
    HNR*~\citep{wang2024lookahead} & & \checkmark & \checkmark & \checkmark & 4.42 & 67.0 & 61.0 & 51.0 & 5.50 & 56.3 & 46.7 \\
    \midrule
    AG-CMTP~\citep{chen2021topological} & & \checkmark & \checkmark & \checkmark & 7.90 & 39.0 & 23.0 & 19.0 & - & - & - \\
    R2R-CMTP~\citep{chen2021topological} & & \checkmark & \checkmark & \checkmark & 7.90 & 38.0 & 26.0 & 22.0 & - & - & - \\
    InstructNav~\citep{long2024instructnav} & & \checkmark & \checkmark & \checkmark & 6.89 & - & 31.0 & 24.0 & - & - & - \\
    LAW~\citep{raychaudhuri2021language} & \checkmark & & \checkmark & \checkmark & 6.83 & 44.0 & 35.0 & 31.0 & 10.90 & 8.0 & 8.0 \\
    CM2~\citep{georgakis2022cross} & \checkmark & & \checkmark & \checkmark & 7.02 & 41.0 & 34.0 & 27.0 & - & - & - \\
    WS-MGMap~\citep{chen2022weakly} & \checkmark & & \checkmark & \checkmark & 6.28 & 47.0 & 38.0 & 34.0 & - & - & - \\
    AO-Planner~\citep{chen2025affordances} & & \checkmark & \checkmark & & 5.55 & 59.0 & 47.0 & 33.0 & 7.06 & 43.3 & 30.5 \\
    Seq2Seq~\citep{krantz2020beyond} & \checkmark & & \checkmark & & 7.77 & 37.0 & 25.0 & 22.0 & 12.10 & 13.9 & 11.9 \\
    CMA~\citep{krantz2020beyond} & \checkmark & & \checkmark & & 7.37 & 40.0 & 32.0 & 30.0 & - & - & - \\
    NaVid~\citep{zhang2024navid} & \checkmark & & & & 5.47 & 49.0 & 37.0 & 35.0 & - & - & -  \\
    Uni-NaVid~\citep{zhang2024uni} & \checkmark & &  & & 5.58 & 53.5 & 47.0 & 42.7 & 6.24 & 48.7 & 40.9  \\
    NaVILA~\citep{cheng2024navila} & \checkmark & & & & 5.22 & 62.5 & 54.0 & 49.0 & 6.77 & 49.3 & 44.0 \\
    StreamVLN~\citep{wei2025streamvln} & \checkmark & & & & 4.98 & 64.2 & 56.9 & 51.9 & 6.22 & 52.9 & 46.0 \\
    InternVLA-N1 (S2)$\dagger$~\citep{wei2025ground} & \checkmark & & & & 4.89 & 60.6 & 55.4 & 52.1 & 6.41 & 49.5 & 41.8 \\
    InternVLA-N1 (S1+S2)$\dagger$~\citep{wei2025ground} & \checkmark & & \checkmark & & 4.83 & 63.3 & 58.2 & 54.0 & 5.91 & 53.5 & 46.1 \\
    NavFoM (Four views) ~\citep{zhang2025embodiednavigationfoundationmodel} &  &\checkmark & & & 4.61 & \textbf{72.1} & 61.7 & 55.3 & 4.74 & 64.4 & 56.2 \\
    \midrule
    \textbf{\MODEL} & &\checkmark& & & \textbf{3.78} & 70.8 & \textbf{66.4} & \textbf{63.9} & \textbf{3.83} & \textbf{69.3} & \textbf{60.0} \\
    \bottomrule
    \end{tabular}
    \end{adjustbox}
    \end{table*}

Following the evaluation standards established in previous methods~\cite{zhang2025embodiednavigationfoundationmodel, wei2025streamvln, cheng2024navila,zhang2024uni, zhang2024navid, long2024instructnav}, we assess the instruction-following capabilities of \MODEL within continuous environments.

\paragraph{Evaluation Setup and Metrics}
We conduct evaluations on the Val-Unseen splits of two distinct datasets of the \textbf{VLN-CE}: \textbf{R2R-CE} \cite{krantz_vlnce_2020}, which focuses on short, goal-oriented commands, and \textbf{RxR-CE} \cite{ku2020room}, which involves longer, temporally grounded path descriptions. We report three standard metrics to measure precision and efficiency: Navigation Error (NE), the average geodesic distance to the target; Success Rate (SR), the percentage of episodes where the agent stops within 3 meters of the goal; and Success weighted by Path Length (SPL), which penalizes inefficient paths.

\paragraph{Results and Analysis}
As shown in Table~\ref{tab:r2r and rxr}, \MODEL demonstrates superior performance across both benchmarks. On the \textbf{R2R-CE} Val-Unseen split, \MODEL achieves an SR of \textbf{66.4\%}, outperforming the previous state-of-the-art method, NavFoM (Four views), by a margin of \textbf{4.7\%}. Notably, we observe a \textbf{8.6\%} improvement in SPL. This significant gain indicates that \MODEL has superior environmental understanding and path planning capabilities, allowing it to navigate more efficiently without compromising success.
A consistent performance trend is observed on the \textbf{RxR-CE} Val-Unseen benchmark, where SR and SPL reach \textbf{69.3\%} and \textbf{60.0\%}, respectively. We attribute these improvements primarily to our model architecture and the synergy derived from joint training on diverse indoor-outdoor, multi-scene, and multi-task datasets.

\subsection{Object-Goal Task}
Following the evaluation standards established in previous methods~\cite{yokoyama2024vlfm, yokoyama2024hm3d, zhang2024uni, zhang2025embodiednavigationfoundationmodel}, we assess the object-goal capabilities of \MODEL on the HM3D-OVON dataset \cite{yokoyama2024hm3d}.

\paragraph{Evaluation Setup and Metrics}
We report results across the three validation splits based on semantic proximity: Val-Seen (categories seen during training), Val-Seen-Synonyms (unseen synonyms of training categories), and Val-Unseen (semantically distinct unseen categories). We report Success Rate (SR) and Success Weighted by Path Length (SPL) as our primary metrics.

\paragraph{Results and Analysis}
As shown in Table~\ref{tab:ovon}, our method outperforms current SOTA approaches using exclusively RGB inputs, eliminating the need for depth sensors or pose. We attribute this success to our proposed high-level cognitive reasoning with low-level motion planning architecture, and the extensive diversity of our training data. Quantitatively, \MODEL surpasses MTU3D~\citep{zhu2025move} by \textbf{13.2\%} in SR on the challenging Val-Unseen split. Furthermore, while MTU3D suffers a significant performance drop of 14.2\% from Val-Seen to Val-Unseen, \MODEL exhibits a negligible decline of only 1.3\%. This minimal generalization gap underscores the consistency of our model's performance across diverse scenarios.

\begin{table*}[t]
  \centering
  \caption{\textbf{Object-Goal Navigation Performance on HM3D-OVON \cite{yokoyama2024hm3d}.} We evaluate \MODEL on the Open-Vocabulary ObjectNav benchmark across three validation splits: Val-Seen, Val-Seen-Synonyms, and Val-Unseen. Our method achieves superior performance using only panoramic RGB input, without requiring depth sensors or odometry.}
  \label{tab:ovon}
  \begin{adjustbox}{max width=\textwidth}
  \begin{tabular}{l|cccc|cc|cc|cc}
    \toprule
    \multirow{2}{*}{\textbf{Method}} &
    \multicolumn{4}{c|}{\textbf{Observation}} &
    \multicolumn{2}{c|}{\textbf{Val-Seen}} &
    \multicolumn{2}{c|}{\textbf{Val-Seen-Synonyms}} &
    \multicolumn{2}{c}{\textbf{Val-Unseen}} \\
    & \textbf{S.RGB} & \textbf{Pano.} & \textbf{Depth} & \textbf{Odo.}
    & \textbf{SR}$\uparrow$ & \textbf{SPL}$\uparrow$
    & \textbf{SR}$\uparrow$ & \textbf{SPL}$\uparrow$
    & \textbf{SR}$\uparrow$ & \textbf{SPL}$\uparrow$ \\
    \midrule
    BC     & \checkmark  &  &  &  & 11.1 & 4.5  & 9.9  & 3.8  & 5.4  & 1.9  \\
    DAgger & \checkmark &  &  &  & 11.1 & 4.5  & 9.9  & 3.8  & 5.4  & 1.9  \\
    RL  & \checkmark &  &  &  & 18.1 & 9.4  & 15.0 & 7.4  & 10.2 & 4.7  \\
    DAgRL   & \checkmark &  &  &  & 41.3 & 21.2 & 29.4 & 14.4 & 18.3 & 7.9  \\
    BCRL    & \checkmark &  &  &  & 39.2 & 18.7 & 27.8 & 11.7 & 18.6 & 7.5  \\
    VLFM~\citep{yokoyama2024vlfm}     & \checkmark &  & \checkmark & \checkmark & 35.2 & 18.6 & 32.4 & 17.3 & 35.2 & 19.6 \\
    DAgRL+OD~\citep{yokoyama2024hm3d}  & \checkmark &  & \checkmark & \checkmark & 38.5 & 21.1 & 39.0 & 21.4 & 37.1 & 19.8 \\
    Uni-NaVid~\citep{zhang2024uni}    & \checkmark &  &  &  & 41.3 & 21.1 & 43.9 & 21.8 & 39.5 & 19.8 \\
    MTU3D~\citep{zhu2025move}         & \checkmark &  & \checkmark & \checkmark & 55.0 & 23.6 & 45.0 & 14.7 & 40.8 & 12.1 \\
    NavFoM (Four views)~\citep{zhang2025embodiednavigationfoundationmodel}         &  & \checkmark &  &  & 40.1 & 27.1 & 45.4 & 32.6 & 45.2 & \textbf{31.9} \\
    \midrule
    \textbf{\MODEL}                  &  & \checkmark &  &  & \textbf{55.3} & \textbf{32.1} & \textbf{55.4} & \textbf{33.2} & \textbf{54.0} & 30.5 \\
    \bottomrule
  \end{tabular}
  \end{adjustbox}
\end{table*}

\subsection{POI-Goal Task}

Adhering to the evaluation framework proposed in our BridgeNav~\cite{zhao2026bridgingindooroutdoorgapvisioncentric}, we assess the performance of \MODEL on the BridgeNav Dataset. This benchmark is specifically designed to test the ``last-meters'' navigation capability, requiring the agent to identify and precisely enter specific POIs from outdoor environments.

\paragraph{Evaluation Setup and Metrics}
The primary objective of this task is to pass through designated entrances, which vary significantly in width across real-world scenes. To rigorously evaluate navigation precision, we report the \textbf{Success Rate (SR)} under three strict distance thresholds to the entrance center: 0.1m, 0.2m, and 0.3m. Furthermore, \textbf{Navigation Efficiency} is measured by calculating the spatial deviation from optimal paths. We report the deviation metrics across average, best-case, and worst-case scenarios to provide a holistic view of path fidelity.

\paragraph{Results and Analysis}
As presented in Table~\ref{poi_main_exp}, \MODEL outperforms existing approaches across all metrics. Most notably, it achieves a substantial \textbf{70.1\%} improvement in SR at the strictest \textbf{0.1m} threshold. This result highlights our model's exceptional fine-grained control capabilities, enabling it to successfully negotiate narrow entrances where baselines often fail. In terms of efficiency, our approach reduces the average trajectory deviation by \textbf{30.5\%} and yields up to a \textbf{55.6\%} improvement in the best-case scenario. These gains suggest that \MODEL's reasoning-driven planning not only finds the target but does so by generating smooth, quasi-optimal paths that closely mirror geometric ground truth.

\begin{table*}[tbp]
\caption{\textbf{POI-Goal navigation evaluation on BridgeNav Dataset \cite{zhao2026bridgingindooroutdoorgapvisioncentric}.} We compare \MODEL with state-of-the-art methods on task success rate under varying distance thresholds (0.1m, 0.2m, 0.3m) and the navigation efficiency. \MODEL demonstrates superior success rate in POI entrance navigation and achieves more efficient path planning.}
\small
\centering
\begin{tabular}{l|ccc|ccc}
 \toprule
\multirow{2}{*}{\textbf{Method}} & \multicolumn{3}{c|}{\textbf{Task Success Rate}} & \multicolumn{3}{c}{\textbf{Navigation Efficiency}} \\ 
    & \textbf{SR (0.1m)} $\uparrow$ & \textbf{SR (0.2m)} $\uparrow$ &  \textbf{SR (0.3m)} $\uparrow$ & \textbf{TR (mean)} $\downarrow$ & \textbf{TR (best)} $\downarrow$ & \textbf{TR (worst)} $\downarrow$ \\ \midrule
NoMaD \cite{sridhar2024nomad} & 4.13 & 15.07 & 29.20 &  31.35 & 5.45 & 85.91 \\
Citywalker \cite{liu2025citywalker} & 13.79 & 41.02 &  65.96 & 15.58 & 0.76 & 56.47 \\
OmniNav \cite{xue2025omninav} & 18.78 & 46.99 & 72.39 & 14.16 & 0.99 & 53.79 \\
\midrule
\textbf{\MODEL} & \textbf{32.14} & \textbf{71.50}  & \textbf{88.68} &\textbf{9.84} & \textbf{0.44} & \textbf{51.38} \\
\bottomrule
\end{tabular}
\label{poi_main_exp}
\end{table*}

\begin{table*}[t]
    \centering
    \caption{\textbf{Quantitative results on the EVT-Bench \cite{wang2025trackvla} (single view).} We evaluate performance using Success Rate (SR), Tracking Rate (TR), and Collision Rate (CR). The symbols $\dagger$ and $\ddagger$ indicate the adoption of GroundingDINO \cite{liu2024grounding} and the SoM \cite{yang2023set}+GPT-4o \cite{hurst2024gpt} pipeline, respectively. Within each category, the top performance is highlighted in \textbf{bold}, and the runner-up is \underline{underlined}.}
    \label{tab:trackingresults}
    \resizebox{\textwidth}{!}{
    \begin{tabular}{l|ccc|ccc|ccc}
    \toprule
    \multirow{2}{*}{\textbf{Method}} & \multicolumn{3}{c}{\textbf{Single-Target Tracking (STT)}} & \multicolumn{3}{c}{\textbf{Distracted Tracking (DT)}} & \multicolumn{3}{c}{\textbf{Ambiguity Tracking (AT)}} \\
    & \textbf{SR$\uparrow$} & \textbf{TR$\uparrow$} & \textbf{CR$\downarrow$} & \textbf{SR$\uparrow$} & \textbf{TR$\uparrow$} & \textbf{CR$\downarrow$} & \textbf{SR$\uparrow$} & \textbf{TR$\uparrow$} & \textbf{CR$\downarrow$} \\
    \midrule
    IBVS$\ddagger$~\citep{gupta2016novel} & 42.9 & 56.2 & 3.75 & 10.6 & 28.4 & 6.14 & 15.2 & 39.5 & \textbf{4.90} \\
    PoliFormer$\dagger$~\citep{zeng2024poliformer} & 4.67 & 15.5 & 40.1 & 2.62 & 13.2 & 44.5 & 3.04 & 15.4 & 41.5 \\
    EVT~\citep{zhong2024empowering} & 24.4 & 39.1 & 42.5 & 3.23 & 11.2 & 47.9 & 17.4 & 21.1 & 45.6 \\
    EVT$\ddagger$~\citep{zhong2024empowering} & 32.5 & 49.9 & 40.5 & 15.7 & 35.7 & 53.3 & 18.3 & 21.0 & 44.9 \\
    Uni-NaVid~\citep{zhang2024uni} & 25.7 & 39.5 & 41.9 & 11.3 & 27.4 & 43.5 & 8.26 & 28.6 & 43.7 \\
    TrackVLA~\cite{wang2025trackvla} & 85.1 & 78.6 & \textbf{1.65} & 57.6 & 63.2 & \underline{5.80} & 50.2 & \underline{63.7} & 17.1 \\
    NavFoM~\citep{zhang2025embodiednavigationfoundationmodel} & 85.0 & 80.5 & - & 61.4 & 68.2 & - & - & - & - \\
    TrackVLA++~\citep{liu2025trackvla++} & \underline{86.0} & \underline{81.0} & \underline{2.10} & \underline{66.5} & \underline{68.8} & \textbf{4.71} & \underline{51.2} & 63.4 & 15.9 \\ \midrule
    \textbf{\MODEL} & \textbf{86.9} & \textbf{87.6} & 8.54 & \textbf{66.7} & \textbf{75.4} & 11.6 & \textbf{67.3} & \textbf{79.5} & \underline{7.05} \\
    \bottomrule
    \end{tabular}
    }
    \end{table*}

\subsection{Person-Following Task}

To assess the model's ability to follow a dynamic person, we evaluate it on the EVT-Bench~\cite{wang2025trackvla}. This benchmark simulates realistic, dynamic social scenarios, requiring the agent not only to detect the target but also to predict their motion under occlusion and distinguish them from distractors.

\paragraph{Evaluation Setup and Metrics}
The evaluation is stratified into three difficulty levels: Single-Target Tracking (STT) (nominal following), Distracted Tracking (DT) (presence of similar-looking pedestrians causing identity confusion), and Ambiguity Tracking (AT) (frequent severe occlusions requiring target re-identification). We report three standard metrics: Success Rate (SR), the percentage of episodes where the agent successfully follows the target for the full duration without losing track; Tracking Rate (TR), the proportion of time steps where the agent maintains the target within a valid distance and field of view; and Collision Rate (CR), measuring navigation safety.

\paragraph{Results and Analysis}
As detailed in Table~\ref{tab:trackingresults}, \MODEL establishes a new state-of-the-art across all categories. In the nominal STT task, it achieves an SR of \textbf{86.9\%}, surpassing the previous best, TrackVLA++, by 0.9\%. The performance gap widens significantly in complex scenarios: in the highly challenging AT task, our model achieves a \textbf{16.1\%} improvement in both SR and TR. This suggests that in complex environments, \MODEL's superior reasoning and target prediction capabilities effectively contribute to safer, more robust tracking planning compared to reactive baselines.

\section{Application}
In this section, we bridge the gap between foundation model capabilities and practical utility by integrating our model into a comprehensive autonomous system. We first introduce the \textbf{Agentic Navigation System} (Sec.~\ref{sec:agentic_nav}), an agentic framework that augments \MODEL with planning, topological memory, and self-reflection to address the heterogeneity of real-world directives. Subsequently, we detail the physical deployment setup (Sec.~\ref{ssec:hardware_setting}) and demonstrations that substantiate the system's operational robustness and capacity in executing complex, long-horizon missions across diverse open-world environments.

\subsection{Agentic Navigation System}
\label{sec:agentic_nav}

\begin{figure*}[t]
    \centering
    \includegraphics[width=1.0\linewidth]{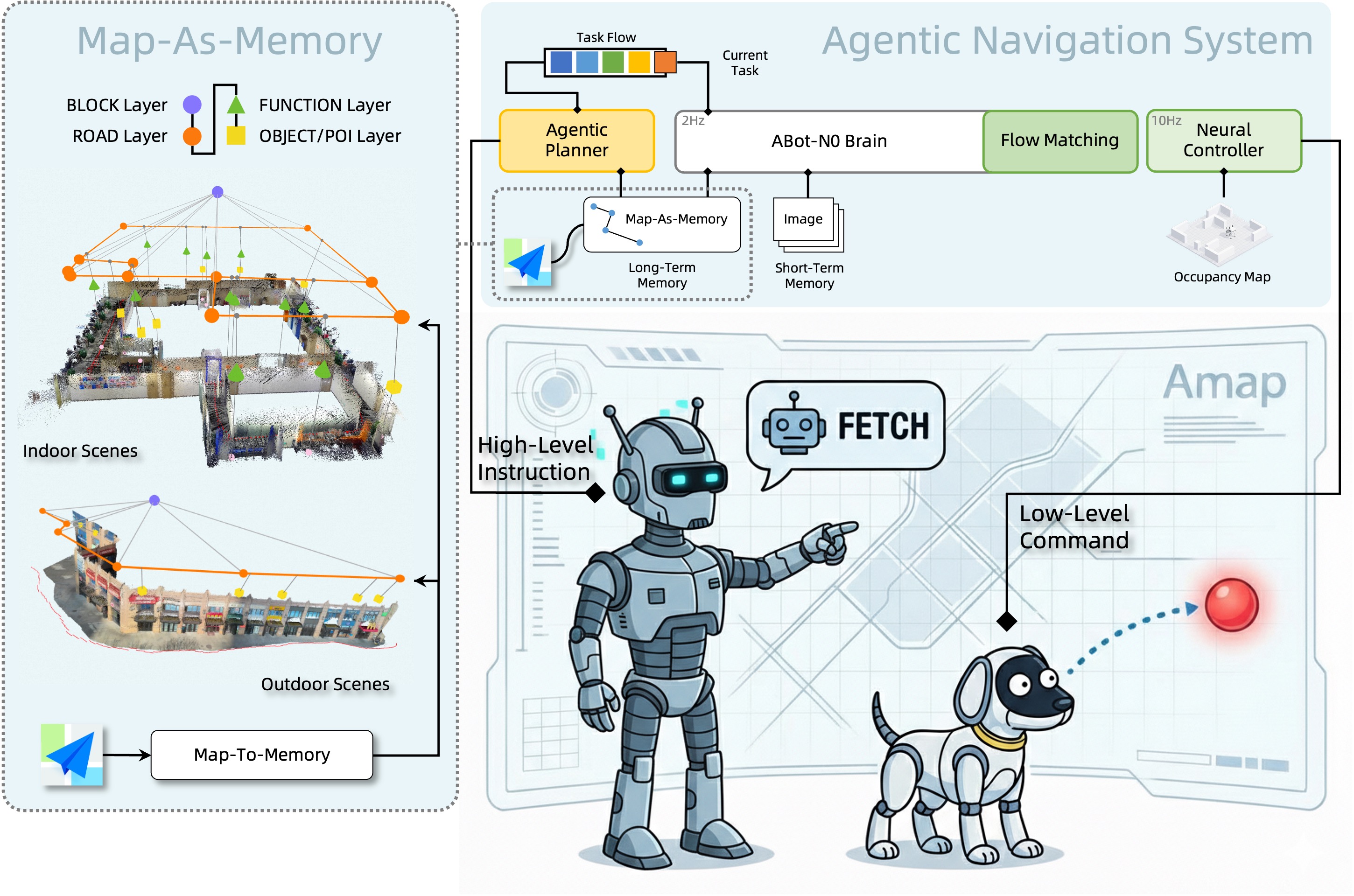}
    \caption{\textbf{Agentic Navigation System Overview.} Our system integrates \MODEL with an agentic framework comprising Agentic Planner, Actor, short-term Episodic Memory, and long-term Topo-Memory to handle complex real-world navigation tasks.}
    \label{fig20}
\end{figure*}

To enable effective deployment of the \MODEL foundation model in complex real-world scenarios, we propose the Agentic Navigation System (Fig. \ref{fig20}). Real-world user instructions are inherently heterogeneous, requiring distinct execution strategies based on the spatial context. A single mission often necessitates a combination of \textit{Approaching} (Point-Goal) for traversing known spaces, \textit{Reaching} (Object-Goal, POI-Goal) for precise target discovery, and \textit{Interaction} (Instruction-Following, Person-Following) for dynamic engagement. 
Recognizing that different navigational phases demand different primitives, our agentic framework systematically decomposes high-level intents into these specific, executable sub-tasks, thereby enabling robust and adaptive long-horizon autonomy.

\subsubsection{Architecture}
Departing from traditional end-to-end control strategies, our system adopts an Agentic VLA framework comprising four core modules: the Agentic Planner, the Actor, the short-term Episodic Memory, and the long-term Topo-Memory. We formulate the task execution process as a Partially Observable Markov Decision Process (POMDP).

Acting as the high-level controller, the Agentic Planner leverages the reasoning capabilities of VLMs to interpret high-level intents. It systematically decomposes these intents into a sequence of sub-tasks executable by the Actor. The Actor module incorporates the \MODEL and a Neural Controller to execute these specific sub-tasks. The short-term Episodic Memory maintains recent observations within the current episode, enabling context-aware decision-making and immediate error recovery. The long-term Topo-Memory serves as a persistent spatial knowledge repository, maintaining hierarchical topological representations that enable cross-scale navigation and experience accumulation. Upon the conclusion of a sub-task, the Self-Reflector of the planner assesses the completion status. In the event of failure, it diagnoses the cause and triggers a re-planning process.

\begin{figure*}[t]
    \centering
    \includegraphics[width=1.0\linewidth]{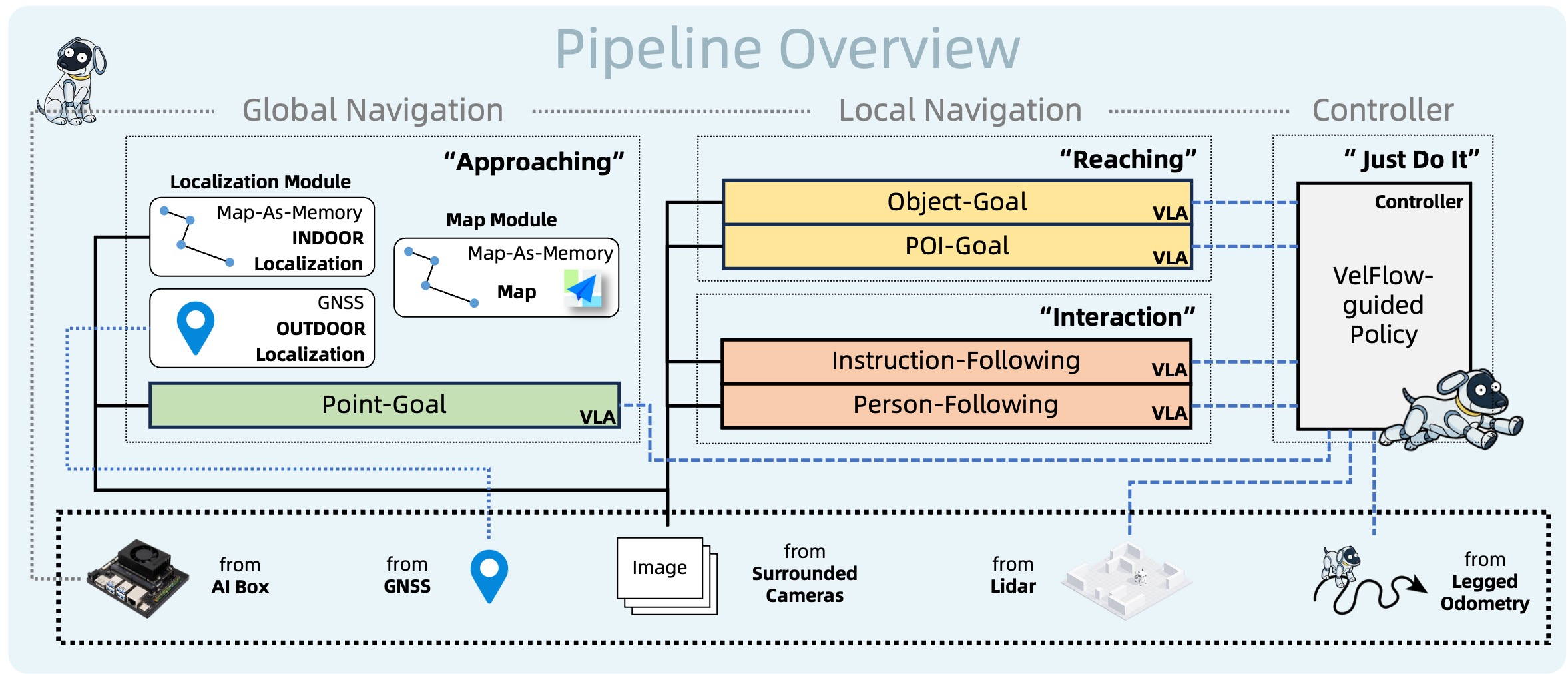}
    \caption{\textbf{Task Execution Pipeline in Agentic Navigation System.} The system orchestrates a hierarchical navigation workflow: (1) Global Navigation employs \textit{Approaching} (Point-Goal) to traverse known spaces using topological memory, (2) Local Navigation utilizes \textit{Reaching} (Object-Goal/POI-Goal) for precise target discovery and \textit{Interaction} (Instruction-Following, Person-Following) for dynamic engagement through visual grounding, and (3) Neural Controller executes low-level velocity-based motion control to realize the planned trajectory.}
    \label{agentsys}
\end{figure*}

\subsubsection{Map as Memory}
To endow Vision-Language Navigation model with human-like environmental familiarity—progressively refining their spatial knowledge across scales ranging from residential interiors to urban environments—we propose Topo-Memory, a spatio-temporal topological memory module. Rather than treating maps as static backdrops, this framework redefines them as a sustained, augmented external memory, enabling the structured deposition and dynamic updating of spatial cognition.

\paragraph{Multi-layer Topology} 
Topo-Memory employs a hierarchical topological graph to organize spatial knowledge. By abstracting environments at multiple granularities, the framework decouples task complexity and ensures cross-scale capabilities from macro-planning to micro-perception:
\begin{itemize}
    \item \textbf{Block Layer:} Defines indoor rooms and outdoor neighborhoods or campus districts. This layer facilitates coarse-grained cross-regional localization and long-range task decomposition.
    \item \textbf{Road Layer:} Characterizes physical connectivity via intersections, doorways, etc., providing rigid reachability constraints for path planning.
    \item \textbf{Function Layer:} Annotates critical semantic nodes such as rest areas, kitchens, and elevator halls, helping to translate abstract linguistic intentions into functional reachability targets.
    \item \textbf{Object/POI Layer:} Identifies specific entities, storefronts, etc. These serve as precise visual-semantic anchors for ``last-meters'' Object/POI-Goal navigation.
\end{itemize}

\paragraph{Unified Indoor-Outdoor Spatial Representation} 
We adopt a ``One Map'' foundation to unify indoor and outdoor navigation. This strategy integrates AMAP-based global routing with real-time visual decision-making. It enables the system to execute complex, multi-stage missions: leaving a home, exiting a residential complex, crossing urban blocks, and finally reaching a specific restaurant inside a shopping mall 3 kilometers away. This creates a comprehensive and robust autonomous navigation system.

\paragraph{Dynamic Maintainability}
Departing from traditional static mapping, Map-as-Memory centers on a ``maintenance-in-the-loop'' mechanism, mitigating cognitive obsolescence caused by dynamic environmental shifts. 
Observations, trajectories, and interaction outcomes from every task execution are back-propagated as evidence into the topological graph. This sustained knowledge deposition enables experience-driven efficiency gains.
In response to environmental dynamics—such as temporary road closures, construction barriers, or corridor congestion—the system adaptively updates graph structures or edge weights. This prevents recursive failures that stem from outdated spatial memory.

\subsubsection{Agentic Planner}
The Planner $\mathcal{P}$ takes the ambiguous user instruction $I$, current observation $O_t$, and historical context $M_{L}$ (Topo-Memory), $M_{S}$ (Episodic Visual Memory) as inputs. Leveraging CoT reasoning, it decomposes the instruction into a sequence of sub-tasks $G = \{g_1, g_2, \dots, g_n\}$ (e.g., Object-Goal, Point-Goal) executable by the \MODEL. This process is formalized as:
$$G = \mathcal{P}(I, M_{L},M_{S},O_t)$$ 
By harnessing the commonsense reasoning capabilities of VLMs, the Planner ensures the following:

\paragraph{Ambiguity Resolution} Free-form natural language instructions are frequently ambiguous and unstructured. VLA often fails to interpret such high-level semantics directly. Acting as the reasoning engine $\mathcal{P}$, our planner leverages the commonsense knowledge embedded within VLMs to parse the ambiguous instruction $I$ and decompose it into a structured, executable sequence of sub-tasks.

\paragraph{Memory-Aware Planning} Our planner integrates a map memory $M_{L}$. During the planning phase, instead of blindly initiating a search, the system first retrieves information from $M_{L}$. For instance, if the instruction is \textit{``find an oven''} and the memory indicates the \textit{``kitchen''} is located at coordinates $(x, y)$, the planner prioritizes generating a deterministic $\text{Point-Goal}(x,y)$ task. This rapidly guides the robot to the target region, avoiding inefficient global Object-Goal exploration. Upon arrival, the system transitions to a local Object-Goal policy to precisely locate the specific target instance.

\paragraph{Strategic "Coarse-to-Fine" Decomposition}
Motivated by the rapid degradation of \textit{Reaching} tasks (Object-goal, POI-goal) performance over long horizons versus the inherent robustness of \textit{Approaching} tasks (Point-Goal), we adopt a ``Coarse-to-Fine'' decomposition strategy. Instead of a single prone-to-failure command, the planner structures the task flow into a reliable sequence: utilizing Point-Goal for the long-range \textit{Approaching} phase, transitioning to Object/POI-Goal for precise local \textit{Reaching}, and concluding with task-specific \textit{Interaction}:
$$P(W | I, M_{L},M_{S}) = \underbrace{P(G | I, M_{L},M_{S})}_{\text{Agentic Planning}} \cdot \prod_{j=1}^{N} \underbrace{P(\mathcal{W}_j | g_j, M_{L},M_{S})}_{\text{\MODEL}}$$

\paragraph{Closed-loop Reflection and Self-Correction}
To bolster system robustness and establish a closed-loop control mechanism, we incorporate a VLM-based Self-Reflector $\mathcal{S}$. Upon the conclusion of a sub-task $g_i$, the Self-Reflector assesses the memory $M_{L},M_{S}$, generating a completion status $r$ and natural language feedback $f$:$$(r, f) = \mathcal{S}(M_{L},M_{S}, g_i)$$In the event of task failure (i.e., $r = \text{False}$), the system leverages the provided feedback $f$ to initiate a re-planning mechanism within the planner, thereby emulating human-like self-correction capabilities:$$G' = \mathcal{P}(I, M_{L},M_{S}, O_t, f)$$

\subsubsection{Neural Controller}

To bridge the latency gap between high-level reasoning and real-time execution, we incorporate a Neural Controller based on the CE-Nav framework~\cite{yang2025nav}. Although \MODEL generates strategic waypoints at a frequency of 2Hz, this rate is insufficient to guarantee stable obstacle avoidance in dense or dynamic environments. The Neural Controller serves as a high-speed reactive layer, operating at over 10Hz on our edge device.

Specifically, we leverage the pre-trained VelFlow expert—a conditional normalizing flow model—as a guiding prior to fine-tune a dynamics-aware refiner via Reinforcement Learning within the Isaac Sim environment. This refinement process is tailored to the specific robot hardware and its underlying locomotion policy. During inference, the Neural Controller takes the latest waypoints provided by the VLA and a real-time occupancy map derived from LiDAR scans as primary inputs. This perception-action loop enables the controller to translate abstract waypoints into precise and dynamically feasible body velocity commands $(v_x, v_y, v_{yaw})$. This hierarchical decoupling ensures both high-fidelity waypoint tracking and robust safety responses during complex missions.

\subsection{Real-world Deployment}

\begin{figure*}[t]
    \centering
    \includegraphics[width=0.5\linewidth]{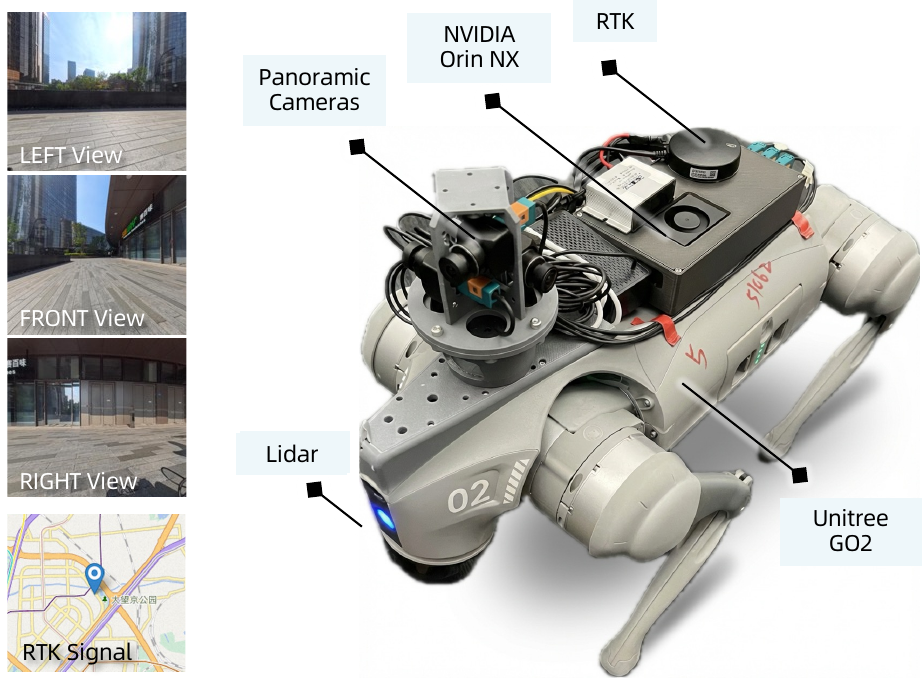}
    \caption{\textbf{Hardware Platform.}
    Our deployment platform is built on a Unitree Go2 X quadrupedal robot equipped with three monocular RGB cameras ($270^\circ$ horizontal FOV), a Unitree 4D LiDAR L2 for occupancy mapping, an RTK-GNSS receiver for global localization, and an onboard NVIDIA Jetson Orin NX module for real-time inference.
    }
    \label{agentsys}
\end{figure*}

To validate the practical applicability of our proposed system, we deploy the complete Agentic Navigation System on a Unitree Go2 quadrupedal robot and conduct extensive real-world experiments in diverse indoor and outdoor environments.

\subsubsection{Hardware setting}
\label{ssec:hardware_setting}

The experimental platform is built upon the Unitree Go2 X quadrupedal robot (Fig.~\ref{agentsys}), a dynamically stable system featuring 12 actuated degrees of freedom. For geometric environmental sensing, the robot is equipped with a Unitree 4D LiDAR L2, providing omnidirectional point cloud data.

\paragraph{Multi-Modal Sensor Suite}
To provide the VLA model with comprehensive environmental context, we integrate a heterogeneous sensor array. Global localization is maintained via a Real-Time Kinematic GNSS (RTK-GNSS) receiver, ensuring geospatial consistency across long-horizon missions. For local scene perception, we deploy three rolling-shutter monocular RGB cameras (FOV: $H120^\circ, V90^\circ$ for each), rigidly mounted to cover the forward, left, and right sectors. This configuration yields a combined horizontal field of view of approximately $270^\circ$, functioning as the ``eyes'' of the robot for precise egocentric navigation and target identification.

\paragraph{Perception and Occupancy Mapping} To bridge the spatiotemporal gap between the low-frequency VLA waypoints planner and the high-frequency velocity controller, we implement a tightly coupled BEV perception framework. This framework fuses proprioceptive legged odometry with a self-developed real-time local mapping pipeline.
Departing from the default height-map provided by the Unitree SDK, we developed a dedicated occupancy grid construction pipeline. By filtering transient obstacle noise and integrating temporal consistency, this pipeline generates a robust occupancy representation. This approach significantly enhances navigation reliability in crowded, human-centered environments while ensuring the state feedback remains spatially grounded for terrain-adaptive execution.

\paragraph{Onboard Computation} All critical perception and execution tasks, including the real-time inference of the optimized VLA model, are performed onboard an NVIDIA Jetson Orin NX module (157 TOPS, 16GB RAM).

\subsubsection{Deployment Strategy}

To balance high-level reasoning with real-time execution, we adopt a hybrid cloud-edge deployment architecture. The Agentic Planner is deployed on a cloud server equipped with an NVIDIA RTX 4090 GPU, leveraging its extensive computational resources for complex intent decomposition and self-reflection. In contrast, the \MODEL and the Neural Controller are deployed locally on an NVIDIA Jetson Orin NX module. This local deployment is critical for ensuring low-latency response and operational safety, enabling the robot to maintain autonomous navigation and obstacle avoidance even in environments with unstable or no network connectivity.

To facilitate the deployment of the VLA on edge hardware, we utilize a lightweight vision encoder and an aggressive visual token compression strategy. Specifically, we employ a compact 93M SigLIP-B/16 as the vision backbone, followed by an MLP projector to map visual features into the language token space. Furthermore, we introduce a token merging mechanism (with a merge size of 4) to compress visual tokens. This design significantly reduces visual forward-pass overhead and shortens the prefill sequence length for the LLM, thereby minimizing computational costs and terminal latency. Ultimately, the edge-side VLA model achieves an onboard inference speed of 2Hz while suffering only a 3\% reduction in performance.

\subsubsection{Deployment Visualization}

\begin{figure*}[t]
    \centering
    \includegraphics[width=1\linewidth]{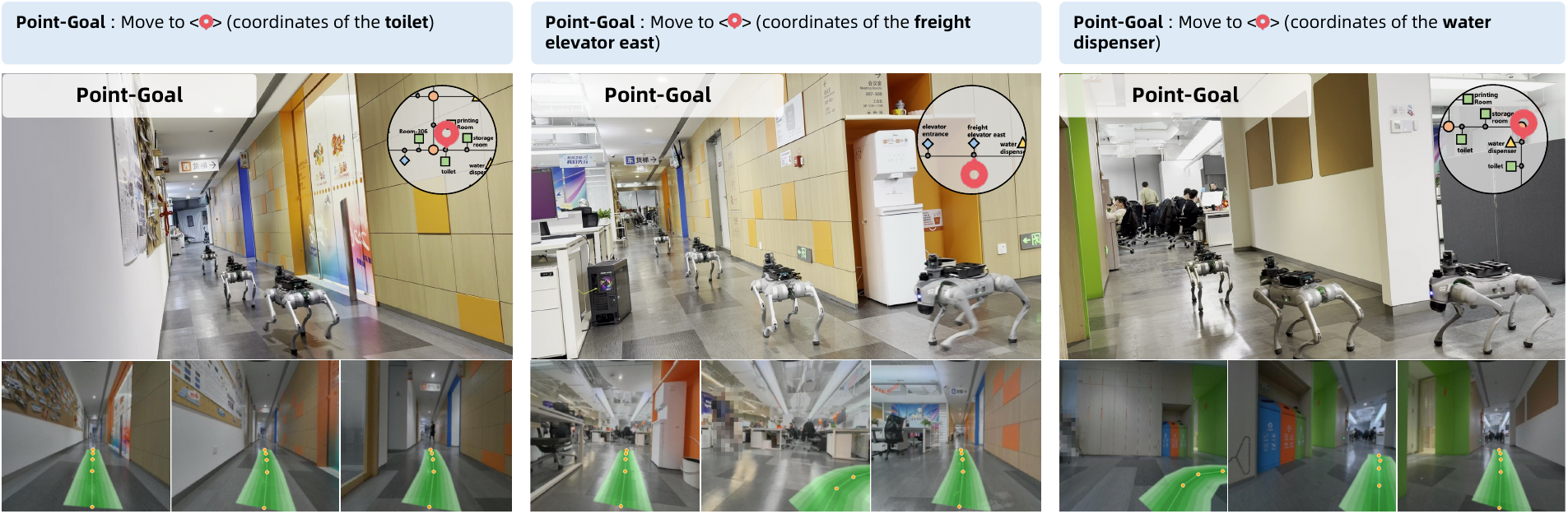}
    \caption{\textbf{Visualization of Point-Goal in real-world deployment.} 
}
    \label{fig:point_goal}
\end{figure*}

\begin{figure*}[t]
    \centering
    \includegraphics[width=1\linewidth]{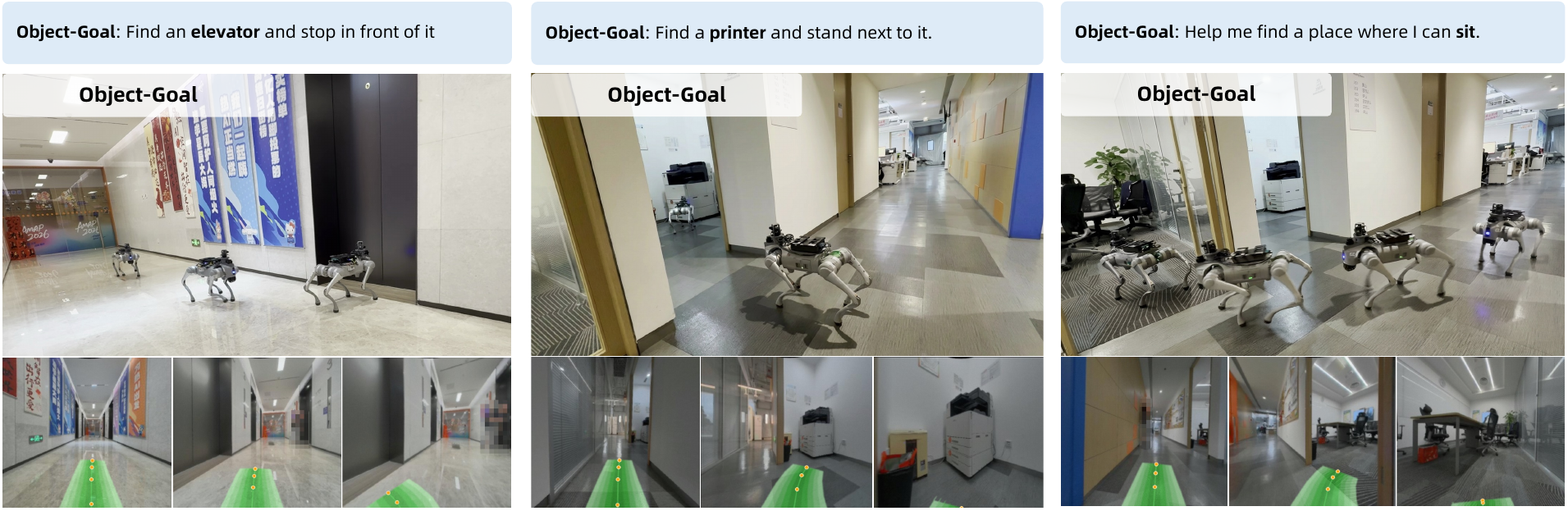}
    \caption{\textbf{Visualization of Object-Goal in real-world deployment.} 
}
    \label{fig:obj_goal}
\end{figure*}

\begin{figure*}[t]
    \centering
    \includegraphics[width=1\linewidth]{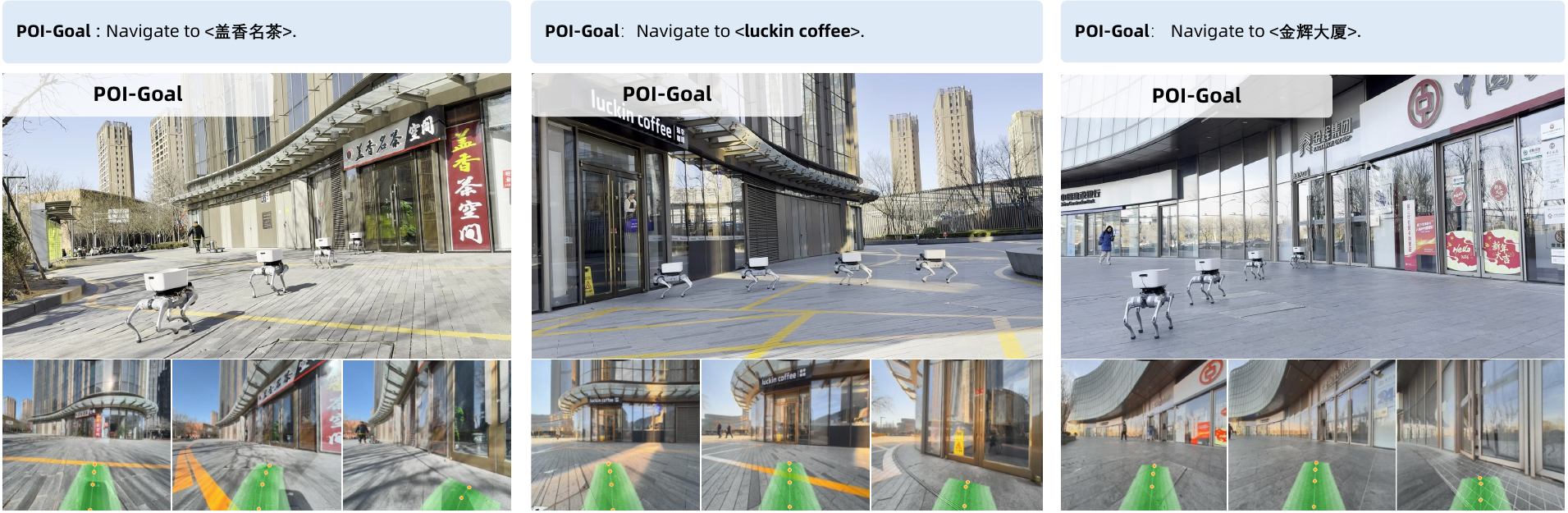}
    \caption{\textbf{Visualization of POI-Goal in real-world deployment.} 
}
    \label{fig:poi_goal}
\end{figure*}

\begin{figure*}[t]
    \centering
    \includegraphics[width=1\linewidth]{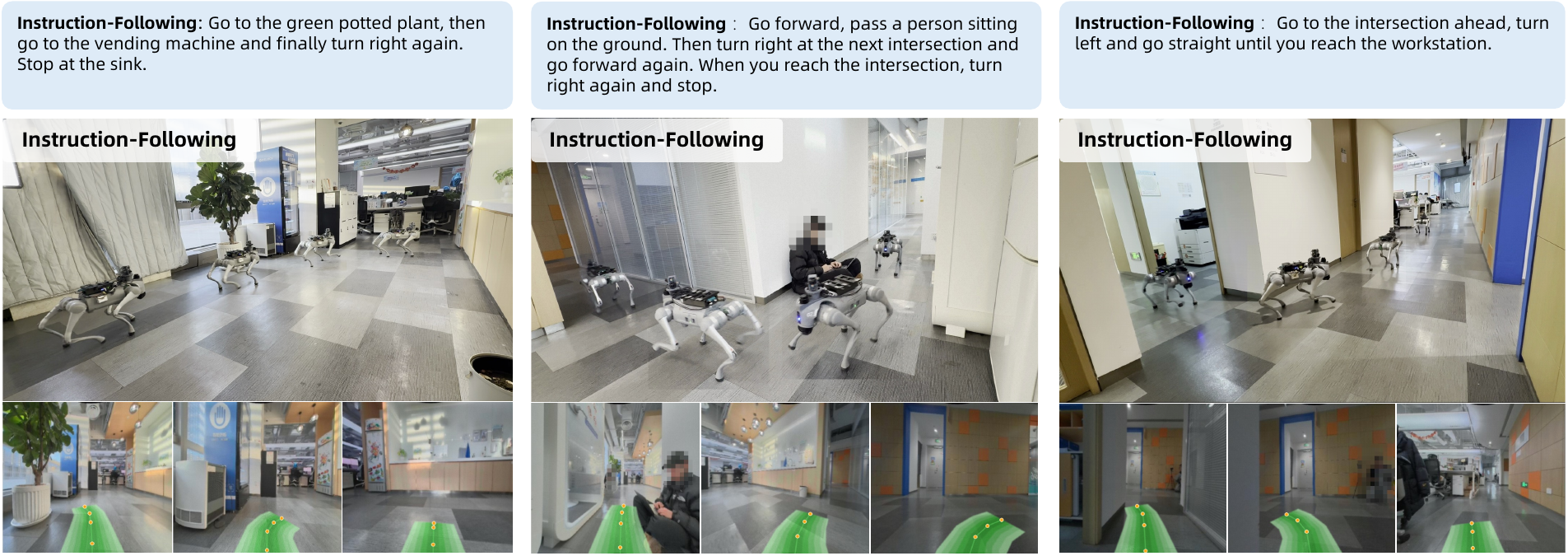}
    \caption{\textbf{Visualization of Instruction-Following in real-world deployment.} 
}
    \label{fig:ins_goal}
\end{figure*}

\begin{figure*}[t]
    \centering
    \includegraphics[width=1\linewidth]{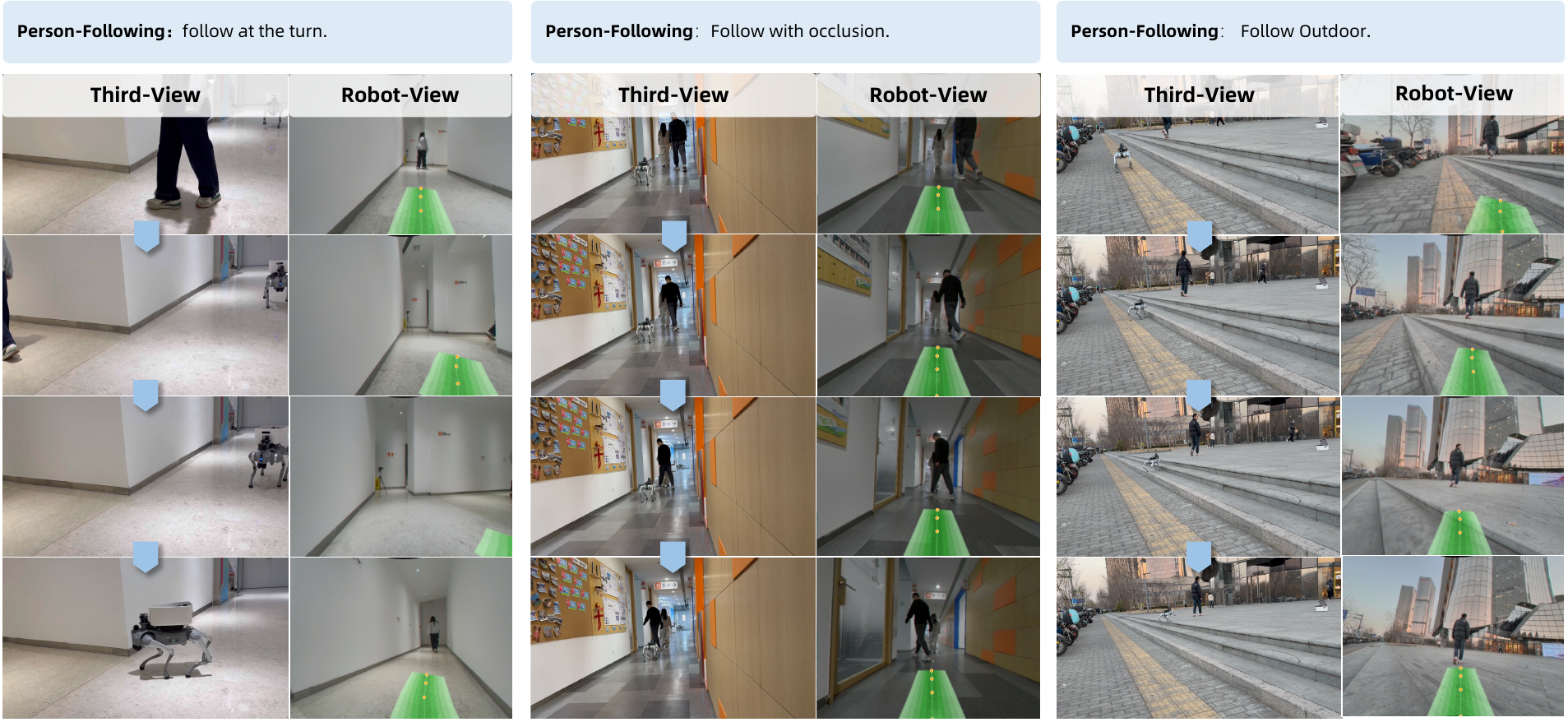}
    \caption{\textbf{Visualization of Person-Following in real-world deployment.} 
}
    \label{fig:person_follow}
\end{figure*}

\begin{figure*}[htbp]
    \centering
    \includegraphics[width=1\linewidth]{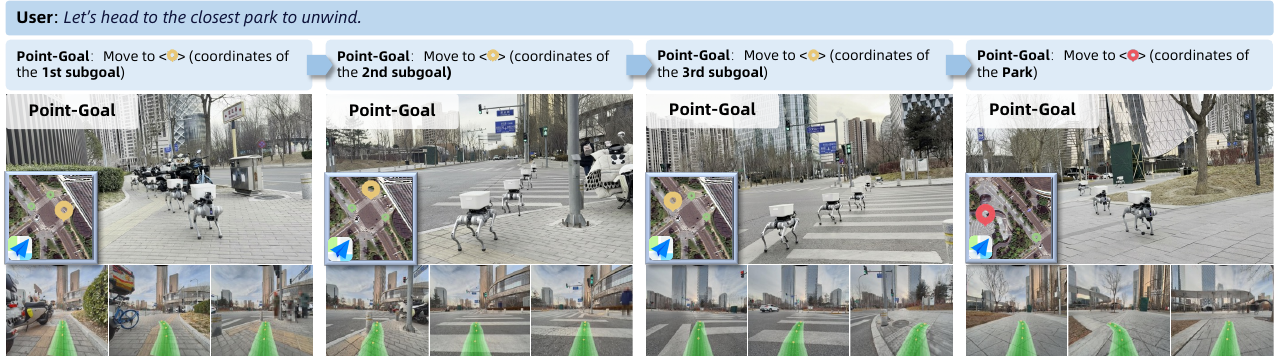}
    \caption{\textbf{Visualization of a long-horizon outdoor navigation task}.
Given the user instruction "Let’s head to the closest park to unwind" the Planner first interprets the user's intent and queries the Topo-Memory to retrieve the target coordinates. Subsequently, it decomposes the global route into a sequence of intermediate sub-goals. These sub-goals are then sequentially executed by the ABot-N0 using its Point-Goal primitive, successfully guiding the robot to the final destination.
    }
    \label{fig:outdoor1}
\end{figure*}

\begin{figure*}[htbp]
    \centering
    \includegraphics[width=1\linewidth]{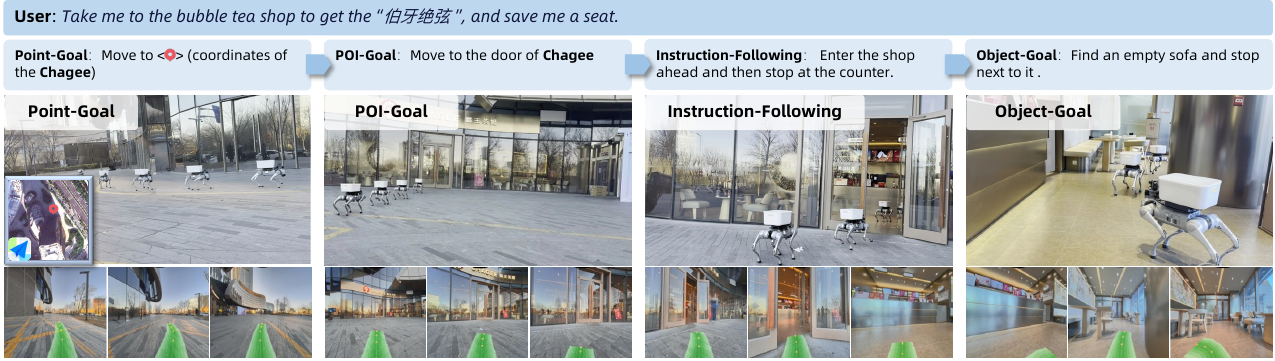}
    \caption{\textbf{Visualization of a complex indoor-outdoor navigation task.} 
Given the user instruction ``Take me to the bubble tea shop to get `Bo Ya Jue Xian' and save me a seat,'' the Planner first queries the Topo-Memory to identify the target POI. Subsequently, it decomposes the mission into a hierarchical sequence of sub-tasks: 
(1) Point-Goal to \textit{Approach} the vicinity of the shop; 
(2) POI-Goal to precisely \textit{Reach} the storefront; 
(3) Instruction-Following to navigate through the door; and 
(4) Object-Goal to locally \textit{Reach} an available sofa. 
The figure depicts the model's predicted trajectory and the corresponding visual observations.
}
    \label{fig:outdoor2}
\end{figure*}

\begin{figure*}[t]
    \centering
    \includegraphics[width=1\linewidth]{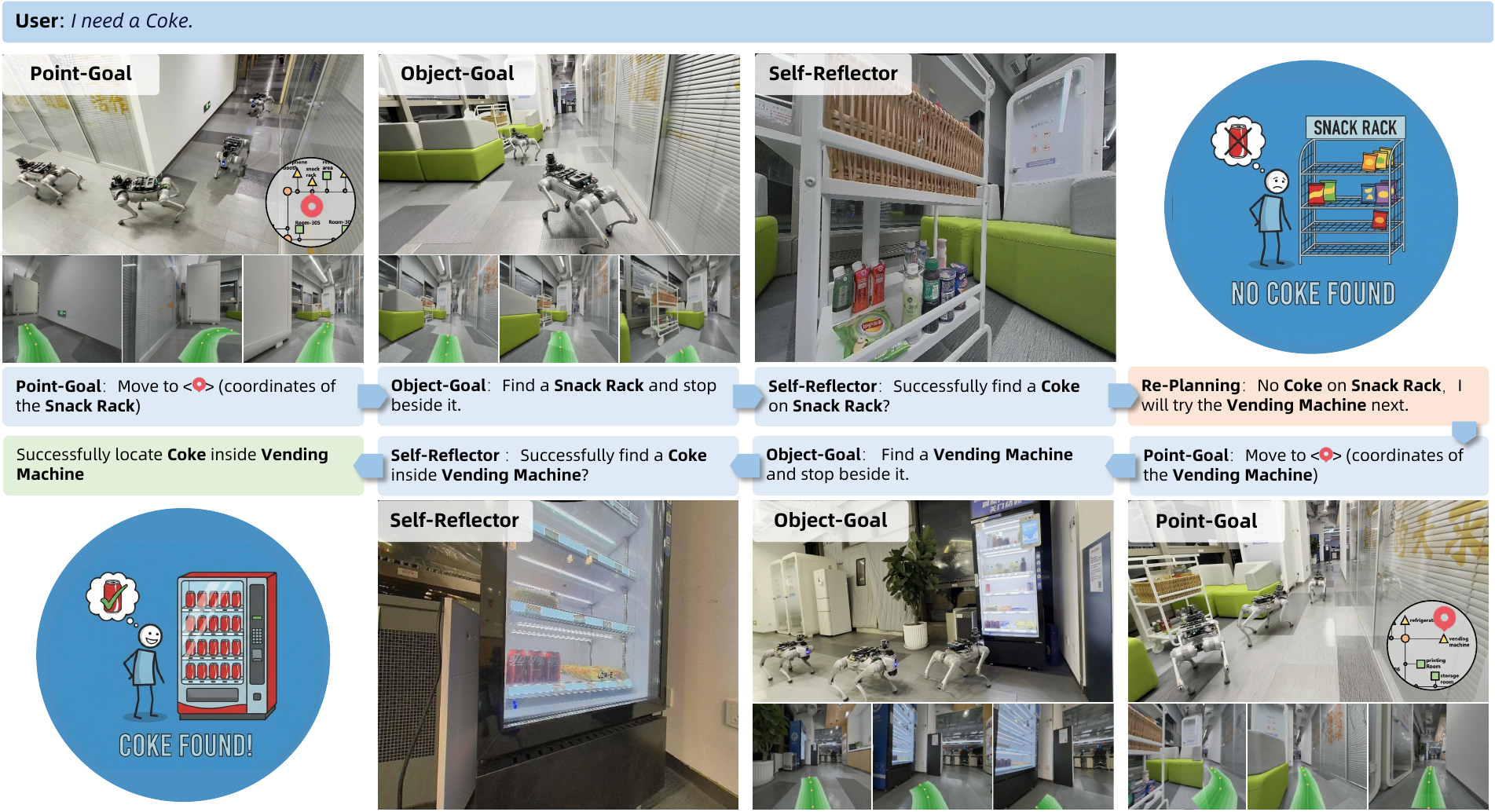}
    \caption{\textbf{Visualization of the Self-Reflector mechanism and adaptive re-planning.} 
Given the user instruction "I need a Coke," the Planner initially queries the Topo-Memory and generates a sequence to reach the \textit{snack rack} (Point-Goal then Object-Goal). 
Upon execution, the Self-Reflector confirms the absence of the target, returning a failure status ($r=\text{False}$) with feedback $f=$ "Coke not found on the snack rack."
Leveraging this feedback, the system triggers a re-planning process to redirect the agent to the \textit{vending machine} (Point-Goal then Object-Goal). 
Finally, the Self-Reflector successfully identifies the Coke in the frontal view ($r=\text{True}$) and terminates the task.
}
    \label{fig:indoor1}
\end{figure*}

\begin{figure*}[t]
    \centering
    \includegraphics[width=1\linewidth]{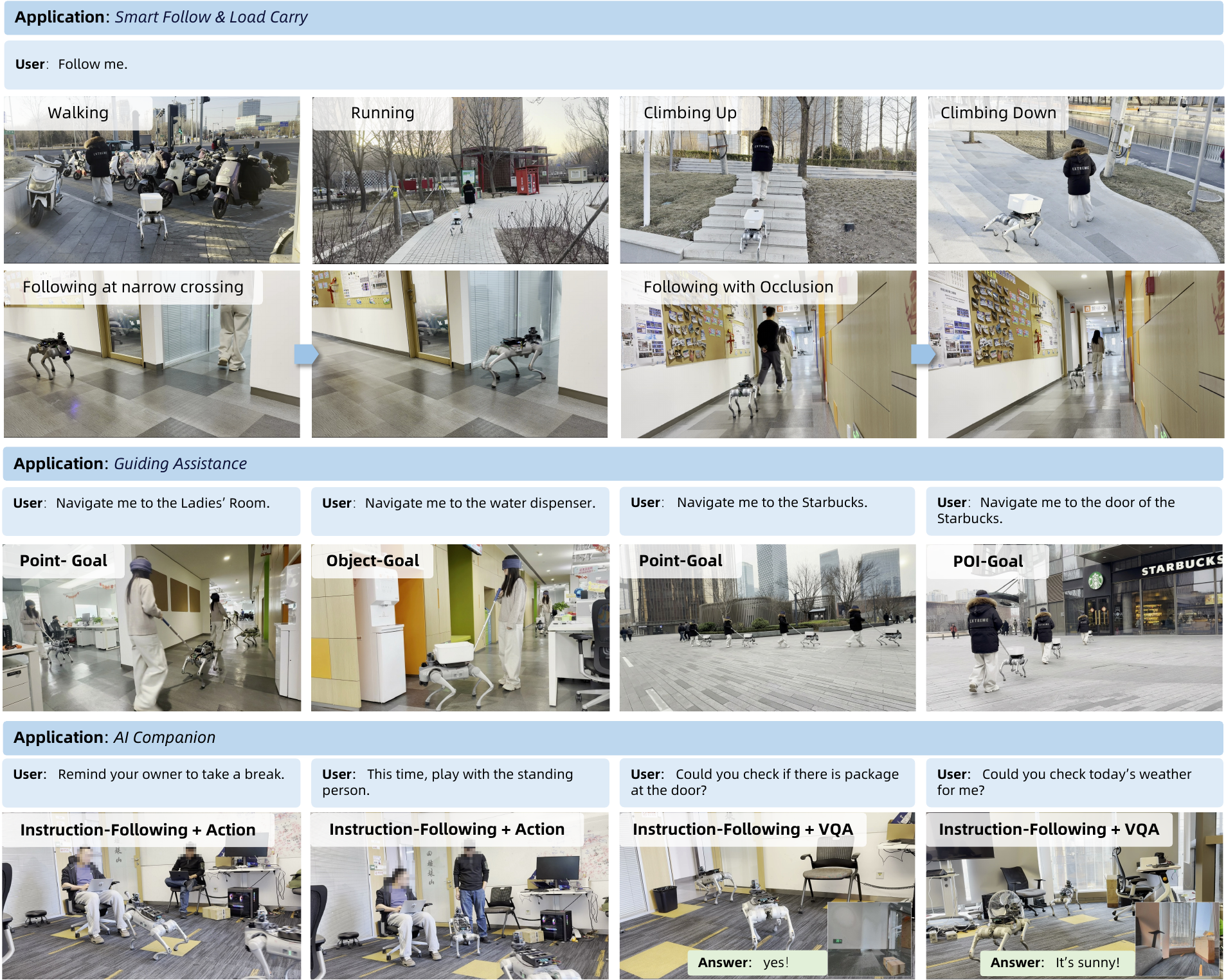}
    \caption{\textbf{Real-world deployment across diverse downstream scenarios.} 
We validate the system's versatility and robustness through three representative downstream applications: \textbf{Smart Follow \& Load Carry}, \textbf{Guiding Assistance} and \textbf{AI Companion} in both indoor and outdoor environments. 
The visualizations demonstrate ABot-N0 adaptability to varying operational requirements.
}
    \label{fig:applications}
\end{figure*}

To comprehensively evaluate the real-world applicability of \MODEL, we design a progressive validation protocol from basic instruction execution to long-horizon complex task handling. We first validate the model's execution precision on core navigation primitives through single-task experiments, including Point-Goal (Fig.~\ref{fig:point_goal}), Object-Goal (Fig.~\ref{fig:obj_goal}), POI-Goal (Fig.~\ref{fig:poi_goal}), Instruction Following (Fig.~\ref{fig:ins_goal}) and Person-Following (Fig.~\ref{fig:person_follow}). We then assess the system's capability to execute long-horizon missions in open-world scenarios. As shown in Fig.~{\ref{fig:outdoor1}--\ref{fig:indoor1}}, these visualizations demonstrate how the system handles comprehensive instructions spanning indoor-outdoor transitions with multi-stage intents. Results indicate that the Agentic architecture enables seamless transitions between navigation phases and exhibits strong closed-loop robustness against environmental uncertainties and task failures. We also demonstrate the system's versatility and feasibility in real-world settings through proof-of-concept experiments across diverse downstream scenarios (Fig.~\ref{fig:applications}).

\section{Conclusion}

In this report, we presented \textbf{\MODEL}, a unified Vision-Language-Action (VLA) foundation model that achieves a ``Grand Unification'' across five core embodied navigation tasks: Point-Goal, Object-Goal, Instruction-Following, POI-Goal, and Person-Following. By adopting a hierarchical ``Brain-Action'' architecture, our model effectively bridges high-level cognitive reasoning with low-level continuous motor control. This synergy is powered by the \MODEL Data Engine, the largest corpus of its kind, comprising 16.9 million expert trajectories and 5.0 million reasoning samples. Comprehensive evaluations demonstrate that ABot-NO sets new state-of-the-art benchmarks across seven authoritative platforms, while real-world deployment of the agentic navigation system on quadrupedal hardware confirms its robustness, generalization and social compliance in dynamic environments.

\clearpage

\section{Contributions and Acknowledgments}
\label{sec:contributions}

\setlength{\parskip}{0pt} 
\setlength{\itemsep}{0pt} 
\setlength{\parsep}{0pt}  
\begin{multicols}{2}
\raggedcolumns

\subsubsection*{Paper Writing}
    \begin{itemize}
        \item Zedong Chu$^{\dagger}$
        \item Shichao Xie$^{\dagger}$
        \item Xiaolong Wu
    \end{itemize}

\subsubsection*{Research}
    \begin{itemize}
        \item Xiaolong Wu
        \item Zedong Chu
        \item Shichao Xie
        \item Yanfen Shen
        \item Minghua Luo
        \item Zhengbo Wang
        \item Fei Liu
        \item Xiaoxu Leng
        \item Junjun Hu
        \item Mingyang Yin
        \item Jia Lu
        \item Yingnan Guo
        \item Kai Yang
        \item Jiawei Han
        \item Xu Chen
        \item Yanqing Zhu
        \item Yuxiang Zhao
        \item Xin Liu
        \item Yirong Yang
        \item Ye He
        \item Jiahang Wang
        \item Yang Cai
        \item Tianlin Zhang
        \item Li Gao
        \item Liu Liu
        \item Mingchao Sun
        \item Fan Jiang
        \item Chiyu Wang
        \item Zhicheng Liu
        \item Hongyu Pan
    \end{itemize}

\columnbreak

\subsubsection*{Data}
    \begin{itemize}
        \item Honglin Han
        \item Zhining Gu
        \item Kuan Yang
        \item Jianfang Zhang
        \item Di Jing
    \end{itemize}

\subsubsection*{Engineering}
    \begin{itemize}
        \item Honglin Han
        \item Zihao Guan
        \item Wei Guo
        \item Guoqing Liu
        \item Di Yang
        \item Xiangpo Yang
    \end{itemize}

\subsubsection*{Hardware}
    \begin{itemize}
        \item Menglin Yang
        \item Hongguang Xing
        \item Weiguo Li
    \end{itemize}

\subsubsection*{Project Leader}
    \begin{itemize}
        \item Mu Xu
        \item Xiaolong Wu
    \end{itemize}

\end{multicols}

\subsubsection*{Acknowledgments}
We extend our sincere gratitude to the broader team for their support, particularly Ziqiao Li, Zixiao Tang and Jingjing Ma.

\vfill
\noindent $^{\dagger}$ Corresponding authors.

\clearpage

\bibliographystyle{plainnat}
\bibliography{main}

\clearpage
\beginappendix

\let\clearpage\relax

\end{document}